\documentclass[11pt]{article}

\usepackage{acl}
\usepackage{times}
\usepackage{latexsym}
\usepackage[T1]{fontenc}
\usepackage[utf8]{inputenc}
\usepackage{microtype}
\usepackage{inconsolata}
\usepackage{graphicx}
\usepackage[table]{xcolor}
\usepackage{amsmath}
\usepackage{amssymb}
\usepackage{booktabs}
\usepackage{multirow}
\usepackage{xspace}
\usepackage{array}
\usepackage{url}
\usepackage{tikz}
\usetikzlibrary{arrows.meta,positioning,fit}

\newcommand{\mathfive}{MATH500\xspace}
\newcommand{\gsm}{GSM8K\xspace}

\newcommand{\sevra}{\textsc{SeVRA}\xspace}
\newcommand{\helpfix}{\textsc{Fix}\xspace}
\newcommand{\harmflip}{\textsc{Flip}\xspace}
\definecolor{sevraBlue}{HTML}{2F6FBB}
\definecolor{sevraGreen}{HTML}{1F8F55}
\definecolor{sevraRed}{HTML}{C9332B}
\definecolor{sevraGold}{HTML}{B7791F}
\definecolor{sevraPurple}{HTML}{7650B5}
\definecolor{sevraInk}{HTML}{25344F}
\definecolor{sevraSoft}{HTML}{F3F6FB}
\newcommand{\best}[1]{\textbf{\textcolor{sevraGreen!70!black}{#1}}}
\newcommand{\risk}[1]{\textcolor{sevraRed!80!black}{#1}}
\newcommand{\costbest}[1]{\textbf{\textcolor{sevraGold!80!black}{#1}}}
\newcommand{\policytag}[1]{\textcolor{sevraInk}{\textsc{#1}}}
\newcommand{\dash}{--}
\newcolumntype{L}[1]{>{\raggedright\arraybackslash}p{#1}}
\newcolumntype{R}[1]{>{\raggedleft\arraybackslash}p{#1}}
\newcommand{\promptbox}[2]{%
\par\smallskip
\noindent\fcolorbox{sevraBlue!45}{sevraSoft}{%
\begin{minipage}{0.96\linewidth}
\textbf{\textcolor{sevraBlue!80!black}{#1}}\par\smallskip
\small #2
\end{minipage}}%
\par\smallskip}

\newcommand{\vtcs}{\textsuperscript{$\heartsuit$}}
\newcommand{\fbri}{\textsuperscript{$\blacklozenge$}}
\newcommand{\fbriCancer}{\textsuperscript{$\lozenge$}}
\newcommand{\corrauth}{\textsuperscript{$\dagger$}}

\title{Think Again or Think Longer?\\
Selective Verification for Budget-Aware Reasoning}

\author{
\begin{tabular}{c}
\textbf{Sajib Acharjee Dip\vtcs\corrauth \quad
Dawei Zhou\vtcs \quad
Liqing Zhang\vtcs\fbri\fbriCancer\corrauth} \\
{\normalfont\mdseries \vtcs Department of Computer Science, Virginia Tech} \\
{\normalfont\mdseries \fbri Fralin Biomedical Research Institute, Virginia Tech} \\
{\normalfont\mdseries \fbriCancer FBRI Cancer Research Center, Washington, DC} \\
{\normalfont\mdseries \corrauth Corresponding author} \\
{\normalfont\mdseries\small \underline{Code:} \url{https://github.com/Sajib-006/SEVRA}} \\
{\normalfont\mdseries\small \underline{Replay dashboard:} \url{https://huggingface.co/spaces/sevra-space/sevra-replay}}
\end{tabular}
}

\begin{document}
\maketitle

\begin{abstract}
Test-time reasoning is increasingly used as a serving-time control knob, but
extra reasoning is not uniformly valuable: it can repair failed attempts, waste
compute on already-correct answers, or introduce harmful answer changes. We
study this as a deployment allocation problem rather than a new-verifier
problem. We introduce \sevra, Selective Verification for Reasoning Allocation,
a serving-layer controller that decides whether to preserve a frozen solver's
initial answer or invoke active verification. Using a frozen Qwen3-4B solver,
we log intervention outcomes and train recoverability-aware gates from
serving-visible attempt state. On \mathfive, selective verification reaches
76.3\% accuracy, compared with 75.5\% for always verifying, while reducing
post-generation tokens by 26.8\% and harmful flips from 2.2\% to 1.0\%.
However, an 8,192-token initial solve reaches 76.0\% accuracy with 28\% fewer
total model tokens, showing that selective recovery is useful but not the best
tested cost frontier. In frozen transfer to \gsm, the selective policy verifies
only 3.0\% of examples, improves accuracy from 93.4\% to 94.5\%, and reduces
verification tokens by 91.2\% relative to always verifying; again, a longer
initial solve matches its accuracy with fewer realized tokens. On
CommonsenseQA, always-on verification hurts, while Self-Consistency@5 improves
accuracy at about five times the realized token cost. The resulting deployment
rule is: tune the initial budget first, then use selective recovery when
explicit checks, bounded retries, auditability, or regression-risk control matter.
\end{abstract}

\section{Introduction}

Inference-time reasoning is increasingly treated as a controllable serving
resource. With more tokens or model calls, a system may continue a solution,
sample alternatives, critique an answer, or actively verify it. These actions
can repair failures, but they also add latency and cost. They can also be
unsafe: a second pass may revise a correct answer into an incorrect one
\citep{huang2024large}.

This creates a deployment question: after observing an initial attempt, should
the system accept it or spend another call? The answer is not determined by
difficulty alone. A difficult problem may already have a correct answer, while
an easier problem may have an incomplete or truncated attempt. What matters is
whether the \emph{current attempt} is recoverable by a specific intervention.

We study \textbf{recoverability-aware selective reasoning} and introduce
\sevra, a lightweight serving-layer controller that decides when a model should
preserve its initial answer or ``think again.'' \sevra uses the problem, base
attempt, and runtime-observable signals to predict whether active verification
is likely to help. Active verification asks the same frozen solver to construct
candidate-specific checks and change the answer only when those checks fail. We
compare this policy against accepting the base answer, continuing the attempt,
always verifying, and allocating a larger initial token budget. The solver and
intervention prompts remain frozen throughout.

Our results show that selective verification is useful, but not a universal
compute optimizer. On MATH500~\citep{hendrycks2021measuring,lightman2024let}, \sevra is the strongest tested
post-generation policy, improving over always verifying while reducing
verification tokens and harmful flips. In frozen transfer to \gsm~\citep{cobbe2021training}, it verifies
only a small fraction of examples, improves over the short-base solver, and
eliminates observed harmful flips. However, on both math benchmarks, a longer
initial solve reaches the same accuracy region with fewer realized model
tokens. Thus, the practical rule is to tune the initial reasoning budget first,
then use selective recovery when explicit verification, bounded retries, or
answer-change auditing matters. A CommonsenseQA~\citep{talmor2019commonsenseqa} diagnostic further shows that
the best inference-scaling action is workload-dependent: always-on
verification hurts, while self-consistency helps only at substantially higher
token cost \citep{wang2022self}.

\paragraph{Contributions.}
We make four contributions:
\begin{enumerate}
\item We formulate post-generation reasoning as an intervention-specific
serving decision, measuring helpful fixes, harmful flips, extra calls, and
realized total model tokens.
\item We provide a budget-matched comparison of accepting, continuing,
actively verifying, always verifying, and increasing the initial reasoning
budget, showing why recovery controllers must be compared against tuned
initial-budget baselines.
\item We show that selective active verification is the strongest tested
post-generation recovery policy, but that longer initial reasoning is more
compute-efficient on the tested math cost frontier.
\item We find that cheap serving-visible execution features nearly match
QLoRA-trained 0.6B and 1.7B gates, making a lightweight feature gate the
most attractive deployment option when its small accuracy gap is acceptable.
\end{enumerate}

\section{Related Work}
\label{sec:related}

\paragraph{Inference-time scaling and allocation.}
Chain-of-thought prompting and self-consistency showed that additional
inference computation can improve reasoning
\citep{wei2022chain,wang2022self}. Subsequent search and
deliberation methods allocate computation across candidate reasoning paths,
including Tree of Thoughts and language-agent tree search
\citep{yao2023tree,zhou2023language}. More recent work studies how to allocate
test-time compute and when longer reasoning is preferable to other forms of
inference scaling \citep{snell2024scaling,muennighoff2025s1}. Closest in
motivation, Solve-then-Learn-style methods formulate allocation as a
cost-sensitive policy-learning problem \citep{zhai2026adaptive}. We share this
cost-aware view, but study a different control point: after observing an
initial attempt, should a serving system spend a second \emph{post-generation}
call to verify or revise it? This makes budget-matched comparison against a
longer initial solve central to our evaluation.

\paragraph{Verification, revision, and self-correction.}
Outcome and process verifiers can guide answer selection, tree search, and
step-level decisions \citep{cobbe2021training,lightman2024let}.
PRM-guided inference uses step-level rewards to navigate reasoning
\citep{ma2023let}, while LATTS and state-level selective verification
adapt verifier effort across intermediate states
\citep{uscidda2025latts,qu2026adaptive}. These methods are often more
fine-grained than our setting, but they require additional verifiers, search
state, or step-level serving control. In parallel, self-refinement and
reflection methods revise outputs using model-generated feedback
\citep{madaan2023self,shinn2023reflexion}, and Socratic
self-refinement decomposes responses into verifiable subquestions before
revision \citep{shi2025ssr}. However, intrinsic self-correction can also fail
or turn correct answers into incorrect ones \citep{huang2024large}. Our
active-verification prompt is intentionally simple: the same frozen solver
constructs candidate-specific checks, while the controller explicitly accounts
for both helpful fixes and harmful flips.

\paragraph{Routing, uncertainty, and frozen-model serving.}
Confidence, semantic uncertainty, and calibration support selective prediction
and help decide when a model output should be trusted
\citep{kadavath2022language,kuhn2023semantic,guo2017calibration}. Cost-aware LLM
routing instead chooses among models or cascades
\citep{chen2023frugalgpt,ong2024routellm}. Recoverability-aware routing has
also been studied in retrieval-heavy QA, where RASER predicts whether to
escalate from a cheap one-shot RAG answer to more expensive multi-hop retrieval
\citep{li2026raser}. SEVRA routes among actions applied to the same frozen
solver after its first attempt, so the key signal is not only uncertainty or
difficulty, but whether the current attempt is recoverable under a specific
intervention. Reasoning-specialized models such as DeepSeek-R1 show that
post-training can produce stronger long-form reasoning
\citep{guo2025deepseek}; our work studies the serving problem that remains
after such a model is chosen: whether to accept, verify, continue, sample, or
allocate a larger initial budget for a particular request.

\section{Problem Formulation}
\label{sec:problem}

For an input problem $x$, a frozen solver produces attempted solution $s_0$,
answer $a_0$, and runtime metadata $m_0$. The metadata includes completion
reason, token counts, finalizer use, and task-level features available during
serving. A controller chooses action $z$ from:
\begin{equation}
z \in \{\textsc{accept},\textsc{continue},\textsc{active-verify}\}.
\end{equation}
Accept returns $a_0$. Continue exposes the existing attempt and asks the solver
to check and revise it. Active verification constructs candidate-specific
checks before preserving or repairing the answer.

Let $c_0 \in \{0,1\}$ denote base correctness and $c_z$ correctness after
action $z$. We define a \textbf{helpful fix} as
\begin{equation}
\helpfix(z)=\mathbb{1}[c_0=0 \land c_z=1],
\end{equation}
and a \textbf{harmful flip} as
\begin{equation}
\harmflip(z)=\mathbb{1}[c_0=1 \land c_z=0].
\end{equation}
A recoverability gate estimates whether active verification will yield a
helpful fix from $(x,s_0,m_0)$, without access to the gold answer.

\paragraph{Cost accounting.}
For each policy, we report accuracy, intervention rate, harmful-flip rate,
action tokens, and total model tokens. Total model tokens include prompt and
generation tokens for the base attempt and all invoked interventions. We
separate configured maximum budgets from realized token use. This distinction
is central: reducing verification calls does not establish overall efficiency
if a different initial allocation reaches the same quality for less total
compute.

\begin{figure*}[ht]
\centering
\includegraphics[width=0.98\textwidth]{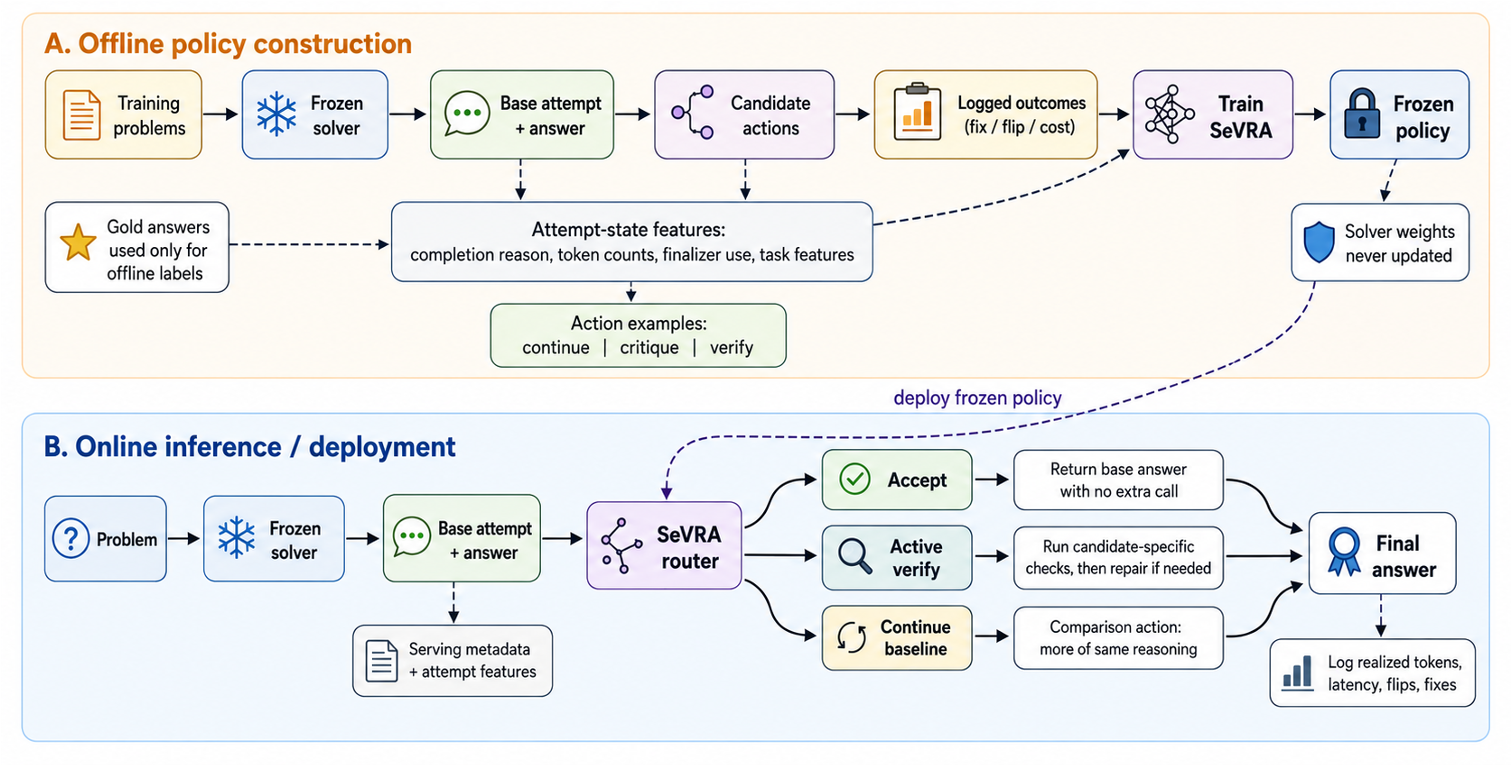}
\caption{Overview of \textsc{SeVRA}. Offline, a frozen solver generates base
attempts, candidate recovery actions are executed, and repair, flip, and token
cost outcomes are logged to train a recoverability-aware policy. At deployment
time, the frozen policy observes only the base attempt and serving metadata,
then routes the example to accept the original answer, actively verify and
repair it, or run a continuation baseline. Realized tokens, extra calls,
helpful fixes, and harmful flips are logged for total-cost evaluation; latency
is discussed as a production replication requirement in
Appendix.}
\label{fig:pipeline}
\end{figure*}

\section{Method}
\label{sec:method}

\subsection{Logged Intervention Outcomes}

We collect attempts from a frozen Qwen3-4B reasoning model
\citep{yang2025qwen3}. For each training example, the model first generates a
base attempt. We then execute candidate post-generation actions and label
whether each action repairs an incorrect answer or damages a correct one. Gold
answers are used only for offline labels and evaluation; they are unavailable
to the deployed gate.

We initially screen continuation, critique-and-repair, and active verification
on 2,000 MATH training examples. Active verification achieves the best static
accuracy and the best fix-to-flip trade-off (Appendix~\ref{app:screening}), so
it becomes the primary selective intervention. This screening step prevents the
controller from being evaluated around an arbitrarily chosen action.

\subsection{Active Verification}

Active verification asks the frozen solver to construct and execute at least
two candidate-specific checks before changing the answer. Checks may reconstruct
the governing equations, test units and bounds, substitute the candidate
answer, or solve through an independent route. The model is instructed to
preserve the original answer if all checks pass and repair it otherwise. The
exact prompt is provided in Appendix~\ref{app:prompts}.

\subsection{Recoverability Gates}

We compare three gate families:
\begin{itemize}
    \item \textbf{Cheap feature gate}: logistic prediction from observable
    task and execution features, including completion status, finalizer use,
    token count, estimated difficulty, verification need, and constraint
    density.
    \item \textbf{Qwen3-0.6B gate}: a 4-bit QLoRA sequence classifier over the
    problem, base attempt, and observable features.
    \item \textbf{Qwen3-1.7B gate}: the same classifier design at 1.7B
    parameters.
\end{itemize}
QLoRA reduces adaptation memory while leaving the base gate weights frozen
\citep{dettmers2023qlora}. Each learned gate predicts whether active
verification produces a helpful fix. Checkpoints are selected by development
AUPRC, but the operating threshold is selected on the held-out MATH
development split by downstream policy accuracy, breaking ties by lower action
tokens. The selected checkpoint and threshold are frozen before \mathfive and
\gsm evaluation.

\begin{table}[t]
\centering
\small
\setlength{\fboxsep}{6pt}
\fcolorbox{sevraBlue!55}{sevraBlue!3}{%
\begin{minipage}{0.94\columnwidth}
\textbf{\textcolor{sevraBlue!85!black}{Algorithm 1: \sevra serving decision}}
\vspace{1mm}

\begin{tabular}{@{}R{0.07\columnwidth}L{0.80\columnwidth}@{}}
\textbf{In} &
Problem $x$, base budget $B_0$, verification budget $B_v$, gate
$g_\theta$, threshold $\tau$. \\
\textbf{1} &
Run frozen solver with budget $B_0$; extract base answer $a^0$ and log
completion reason, finalizer use, and realized tokens. \\
\textbf{2} &
Build serving-visible features from $x$, the base attempt, and runtime
metadata. No gold labels or hidden solver states are available. \\
\textbf{3} &
Compute recoverability score $s=g_\theta(x,b,m)$. \\
\textbf{4} &
If $s\ge\tau$, run active verification with budget $B_v$ and return its
checked answer $a^v$; otherwise return $a^0$. \\
\textbf{5} &
Log answer changes, total realized model tokens, helpful fixes, and harmful
flips for offline monitoring. \\
\end{tabular}
\end{minipage}}
\caption{\sevra as a serving-layer procedure. The controller changes only the
post-generation decision; the solver and verification model remain frozen.}
\label{tab:sevra-algorithm}
\end{table}


\section{Experimental Setup}
\label{sec:setup}

\paragraph{Models and benchmarks.}
We use Qwen3-4B through Ollama as the frozen solver and intervention model
\citep{yang2025qwen3}; solver weights are never updated. Gate training uses
Qwen3-0.6B and Qwen3-1.7B with 4-bit QLoRA. We construct recovery data from
2,000 MATH training examples \citep{hendrycks2021measuring}, split 80/20 by example
for gate training and threshold selection. Evaluation uses the untouched
500-example \mathfive test set and all 1,319 \gsm test examples
\citep{cobbe2021training}. MATH-trained gates and thresholds transfer to
\gsm without fine-tuning or recalibration.

\paragraph{Budgets and evaluation.}
The short-base solver receives a 4,096-token generation limit. Continuation and
active verification each receive up to 4,096 additional generation tokens when
invoked, while the long-base baseline receives an 8,192-token initial budget
and no post-generation action. If a reasoning call does not expose a final
answer, we use a non-reasoning finalizer with a 512-token limit. We report
realized prompt-plus-generation model tokens, including finalizer calls, rather
than configured maximums. Final answers are scored with exact matching and
mathematical equivalence checking. The main selective row uses the 1.7B gate because it obtains the highest
\mathfive accuracy, but we highlight the cheap-feature gate as the practical
deployment default when avoiding an additional served classifier is more
important than a 0.4-point MATH500 gain. We also report the 0.6B gate and
simple heuristic routing baselines. Confidence intervals and significance tests use
paired bootstrap resampling over evaluation examples.


\section{Results and Analysis}
\label{sec:results}
\label{sec:analysis}

\subsection{Budget-Matched Results}

\begin{table*}[ht]
\centering
\small
\setlength{\tabcolsep}{3pt}
\resizebox{\textwidth}{!}{
\begin{tabular}{@{}L{0.15\textwidth}L{0.21\textwidth}R{0.05\textwidth}R{0.07\textwidth}R{0.07\textwidth}R{0.05\textwidth}L{0.38\textwidth}@{}}
\toprule
\textbf{Dataset} & \textbf{Policy} & \textbf{Acc.} &
\textbf{Extra calls} & \textbf{Total tok.} & \textbf{Flips} &
\textbf{Operational reading} \\
\midrule
\rowcolor{gray!7}
\mathfive & Base, 4,096 limit & 59.0 & 0.0\% & 4,313 & \costbest{0.0} &
Many failures are truncation-related. \\
\mathfive & Always continue & 72.0 & 100.0\% & 8,007 & \risk{3.6} &
Repairs failures, but flips correct answers. \\
\mathfive & Selective continue & 73.6 & 48.2\% & 7,064 & \risk{1.7} &
Improves continuation at half the calls. \\
\mathfive & Always active verify & 75.5 & 100.0\% & 8,125 & \risk{2.2} &
Strong recovery, but costly and flip-prone. \\
\rowcolor{sevraGreen!7}
\mathfive & \policytag{Selective active verify} & \best{76.3} & 48.2\% &
7,104 & \risk{1.0} & Best post-generation result; fewer flips and action tokens. \\
\rowcolor{sevraGold!10}
\mathfive & Long base, 8,192 limit & 76.0 & 0.0\% & \costbest{5,124} &
\costbest{0.0} & Best tested cost frontier; no second call. \\
\midrule
\rowcolor{gray!7}
\gsm & Base, 4,096 limit & 93.40 & 0.0\% & 1,180 & \costbest{0.00} &
High base accuracy; little room to verify. \\
\gsm & Always active verify & 93.40 & 100.0\% & 2,932 & \risk{1.25} &
No accuracy gain; adds flips and tokens. \\
\rowcolor{sevraGreen!7}
\gsm & \policytag{Selective active verify} & 94.47 & 3.0\% & 1,335 &
\costbest{0.00} & Sparse recovery; no observed flips. \\
\rowcolor{sevraGold!10}
\gsm & Long base, 8,192 limit & \best{94.54} & 0.0\% & \costbest{1,157} &
\costbest{0.00} & Same accuracy region; fewer realized tokens. \\
\midrule
\rowcolor{gray!7}
CommonsenseQA & Base, 4,096 limit & 76.49 & 1 call & \costbest{2,234} &
\costbest{0.00} & Short-answer workload; different regime. \\
CommonsenseQA & Always active verify & 72.32 & 2 calls & 4,794 &
\risk{5.94} & Verification hurts under workload shift. \\
\rowcolor{sevraPurple!8}
CommonsenseQA & Self-Consistency@5 & \best{78.38} & 5 calls & 11,343 &
\risk{1.56} & Sampling helps, but costs five calls. \\
CommonsenseQA & SC sampled-rollout oracle & 85.18 & \dash & \dash & \dash &
Selection headroom; not deployable. \\
\bottomrule
\end{tabular}}
\caption{Unified main results across the math benchmarks and CommonsenseQA.
Accuracy, extra-call rate, and harmful flips are percentages unless shown as
call counts. Total tokens are realized prompt-plus-generation model tokens
averaged over all examples. Green shading marks the best post-generation
policy; gold shading marks the strongest tested cost frontier; purple marks a
published multi-sample baseline.}
\label{tab:main-results}
\label{tab:math500}
\label{tab:gsm8k}
\label{tab:csqa-main}
\end{table*}

Table~\ref{tab:main-results} summarizes accuracy, extra calls, realized tokens,
and harmful flips across all workloads. On \mathfive, selective active
verification is the strongest tested post-generation policy: it improves over
selective continuation by 2.7 accuracy points under nearly identical
total-token budgets, and compared with always verifying it is 0.8 points more
accurate, reduces action tokens by 26.8\%, and lowers harmful flips from 2.2\%
to 1.0\%. The accuracy difference over always verifying is not statistically
significant ($p=.103$), but the flip reduction is significant ($p<.001$).
However, the long-base baseline reaches 76.0\% accuracy, statistically
comparable to selective verification, while using 28\% fewer total tokens and
no post-generation call. Thus, selectivity improves verification, but does not
beat a better initial allocation on the tested math cost frontier.

The same pattern appears in frozen transfer to \gsm. Selective verification
checks only 3.0\% of examples, improves accuracy by 1.06 points over always
verifying (95\% CI $[0.53,1.63]$, $p<.001$), reduces verification-action tokens
by 91.2\%, and eliminates observed harmful flips. Yet the long-base policy
again reaches statistically indistinguishable accuracy (selective minus long:
$-0.08$ points, 95\% CI $[-0.91,0.80]$, $p=.899$) with 178 fewer realized
tokens per example. The larger configured budget reduces truncation and
finalizer overhead enough to lower realized cost, showing why maximum-token
settings are capacity limits rather than direct cost measures.


\begin{table}[ht]
\centering
\small
\setlength{\tabcolsep}{4pt}
\resizebox{\columnwidth}{!}{
\begin{tabular}{@{}lrrrr@{}}
\toprule
\multirow{2}{*}{\textbf{Gate}} &
\multicolumn{2}{c}{\textbf{MATH500}} &
\multicolumn{2}{c}{\textbf{GSM8K}} \\
& \textbf{Acc.} & \textbf{Verify} & \textbf{Acc.} & \textbf{Verify} \\
\midrule
\rowcolor{sevraGold!10} Cheap features & 75.9 & \costbest{45.0} & 94.47 & \costbest{2.8} \\
Qwen3-0.6B QLoRA & 75.7 & 46.4 & 94.47 & 3.0 \\
\rowcolor{sevraGreen!7} Qwen3-1.7B QLoRA & \best{76.3} & 48.2 & 94.47 & 3.0 \\
\bottomrule
\end{tabular}}
\caption{Frozen gate comparison. Accuracy and verification rate are
percentages. The cheap gate is operationally attractive because it avoids
serving an additional language model.}
\label{tab:gates}
\end{table}

\vspace{-0.6em}
\subsection{Cost Frontier and Workload Shift}

\begin{figure*}[t]
\centering
\includegraphics[width=0.98\textwidth]{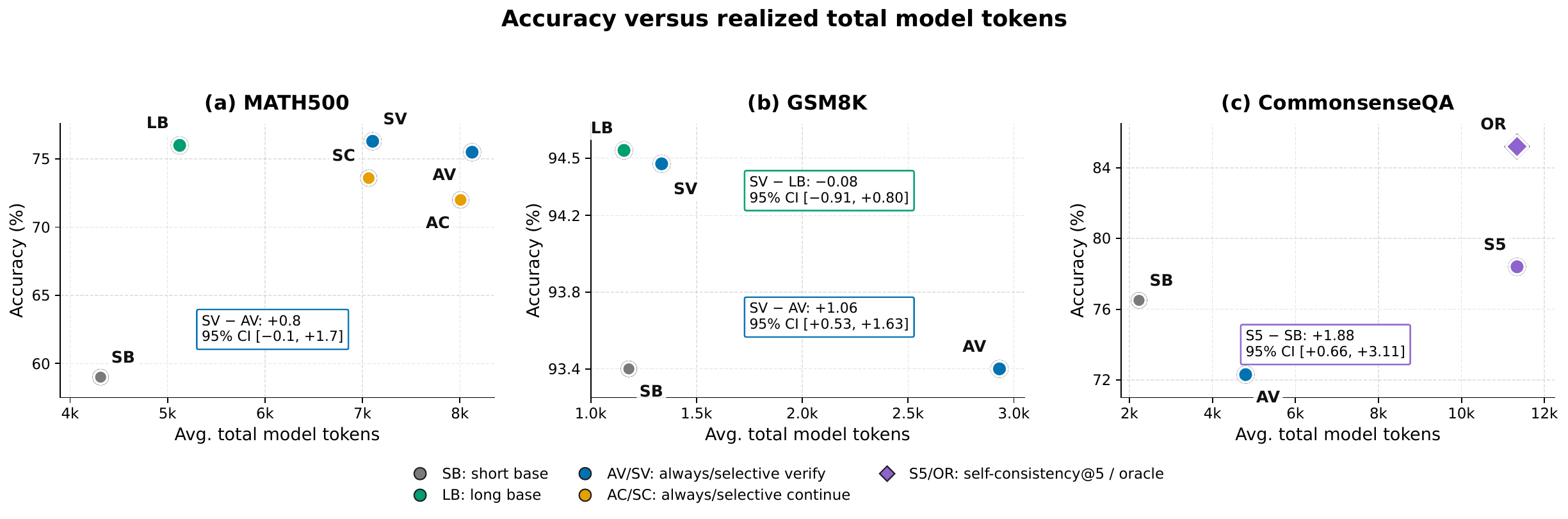}
\caption{Accuracy versus realized total model tokens across MATH500, GSM8K,
and CommonsenseQA. Boxed intervals show paired bootstrap confidence intervals.
Token counts are realized total model tokens rather than configured maximum
budgets.}
\label{fig:accuracy-token-frontier}
\end{figure*}

Figure~\ref{fig:accuracy-token-frontier} visualizes the cost--accuracy
trade-off. On the math benchmarks, selective verification is the best
post-generation recovery policy, while the longer initial solve lies on the
best tested cost frontier. CommonsenseQA shows a different regime: active
verification is the wrong default action, lowering accuracy by 4.17 points and
creating harmful flips. Self-Consistency@5 improves over the base solver by
1.88 points (95\% CI $[0.66,3.11]$, $p=.003$), but uses roughly five times the
realized model tokens. The sampled-rollout oracle indicates that useful answers
often exist among additional samples, but a deployable system still needs a
reliable selection mechanism.

\subsection{Gate Complexity}

Table~\ref{tab:gates} shows that cheap serving-visible features are competitive
with learned gates. On \gsm, all three gates obtain 94.47\% accuracy at about a
3\% verification rate. On \mathfive, the 1.7B gate leads the cheap feature gate
by only 0.4 points. Although the learned gates reach development AUROC near
0.957, this ranking quality gives little downstream advantage over simple
execution features. For deployment, the cheap gate is attractive because it
avoids serving an additional language model and reduces latency, memory, and
maintenance overhead.

\subsection{Fixes, Flips, and Attempt State}

\begin{figure}[t]
\centering
\includegraphics[width=0.99\columnwidth]{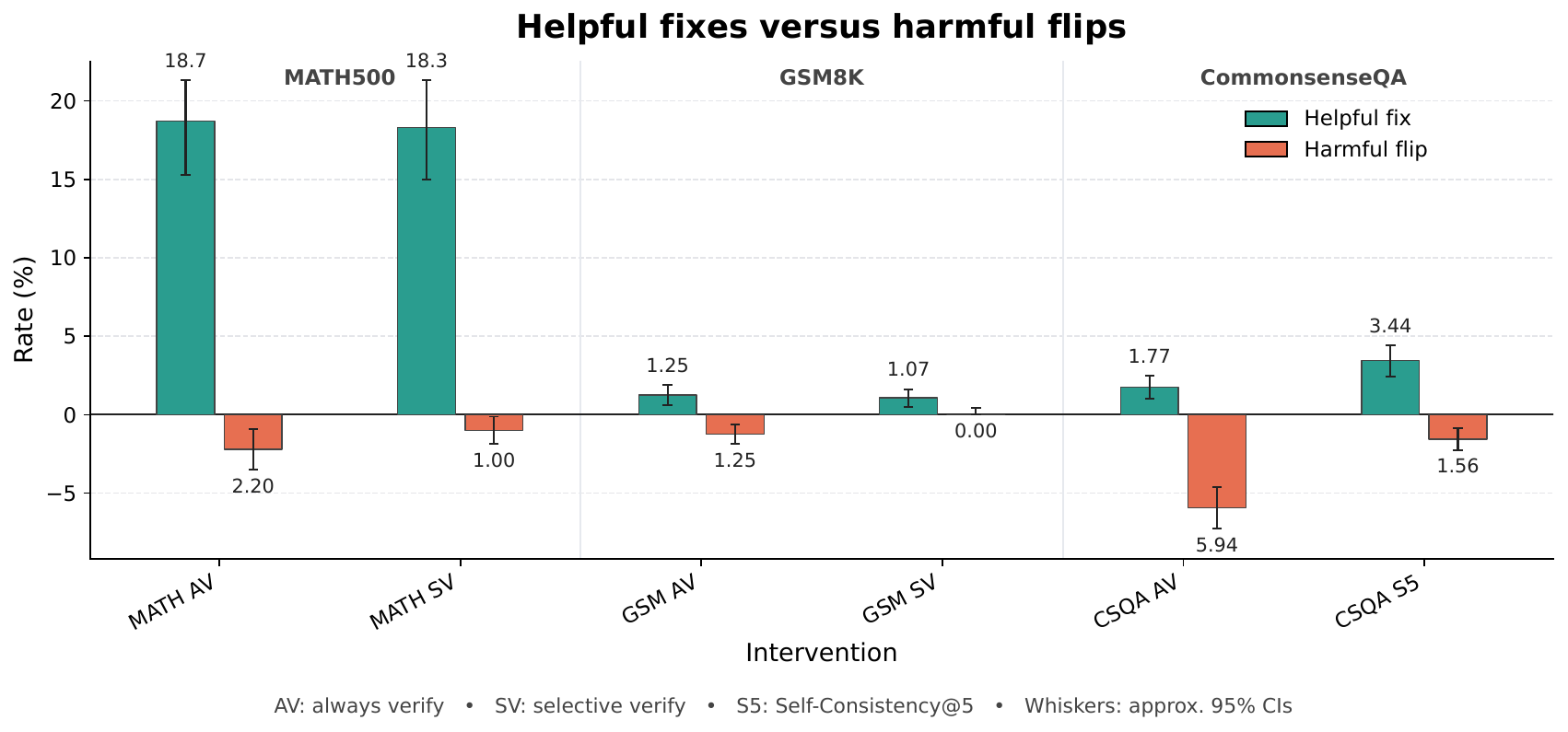}
\caption{Helpful fixes and harmful flips for the main post-generation
interventions. Extra reasoning is useful only when repairs outweigh regressions:
selective verification preserves most MATH500 fixes while reducing flips,
GSM8K has a small recoverable subset, and CommonsenseQA shows that always-on
verification can be unsafe. Whiskers show approximate 95\% binomial intervals;
detailed normal-stop and length-stop subgroup results are in
Appendix~\ref{app:subgroups}.}
\label{fig:fix-flip}
\end{figure}

Figure~\ref{fig:fix-flip} explains the reliability trade-off behind the
aggregate results. On \mathfive, active verification produces many helpful
fixes but also harmful flips; selective verification preserves most fixes while
reducing regressions. On \gsm, both fixes and flips are rare, so a small
selective subset is sufficient. On CommonsenseQA, always verifying is unsafe:
harmful flips outnumber helpful fixes.

Attempt state explains much of the transfer behavior. Only 38 of 1,319
short-base \gsm attempts are truncated; their base accuracy is 15.8\%, and
verification raises it to 52.6\%. Among the 1,281 completed attempts, base
accuracy is 95.7\%; always verifying lowers it to 94.6\%, while selective
verification preserves 95.7\%. This also clarifies why intervention type
matters. For truncated \mathfive attempts, active verification reaches 48.7\%
accuracy, compared with 43.0\% for continuation; for completed attempts,
unnecessary post-generation calls can damage correct answers. More reasoning is
therefore not a single action: verification, continuation, and longer
initial solving expose different cost and regression profiles.

\section{Industry Implications}
\label{sec:industry}

Our results suggest a compact deployment rule: tune the initial reasoning
budget before adding a recovery controller, then use selective verification
when explicit checks, bounded retries, auditability, or regression control are
operationally important. Completion reason, token count, and finalizer use are
cheap serving-visible signals, and our results show that such features nearly
match learned gates. This makes the cheap feature gate the most practical
default controller in our tested setting. At the same time, answer changes
should be treated as reliability risks rather than only as aggregate accuracy
changes: helpful fixes and harmful flips should be monitored separately, with
thresholds chosen according to the application's tolerance for regressions.
Finally, deployment evaluation should measure the whole serving path: base
attempts, intervention prompts, finalizers, retries, extra calls, realized total
tokens, and, in production replications, p50/p95/p99 latency. Because our logs
do not contain per-request wall-clock timings, we use realized model tokens as
the reproducible cost metric and provide the latency-accounting protocol in
Appendix~\ref{app:industry} which gives a
detailed deployment playbook and describes our static replay dashboard.

\section{Conclusion}

Additional reasoning should be treated as an intervention with uncertain
value, not a universally beneficial extension of inference. Recoverability-aware
selection substantially reduces unnecessary verification and harmful answer
changes, while active verification is more effective than simply continuing an
attempt. However, longer initial reasoning is the most compute-efficient tested
strategy on both benchmarks. Practical systems should therefore first choose
an appropriate initial budget, then use selective recovery when explicit
verification or retry behavior is operationally needed. 

\clearpage
\section*{Limitations}

Our experiments use one solver family and public benchmark workloads rather
than production traffic. MATH and GSM8K are controlled stress tests for
reasoning, truncation, and answer extraction, while CommonsenseQA provides only
a lightweight non-mathematical diagnostic. We therefore do not claim deployment
validation in a live product environment. Recoverability is strongly related to
length-limit termination under the tested short-budget configuration, and other
serving stacks may expose different failure modes. Logged sampled rollouts
provide noisy intervention-value labels. Exact matching and symbolic
equivalence do not capture all answer-quality dimensions. Token count and call count are incomplete cost proxies because they omit
wall-clock latency, memory pressure, batching effects, energy use, and provider
pricing. Our logs did not record per-request wall-clock latency or provider
prices, so the main results report realized prompt-plus-generation model tokens
rather than measured p50/p95 serving latency or dollar cost. Appendix gives the latency fields and token-price sensitivity
analysis needed for a production replication. The learned gates are trained on only 2,000 MATH
examples, and transfer is evaluated on \gsm and CommonsenseQA without
retraining. We do not claim that the controller detects general reasoning
failures or that selective verification is more efficient than a well-tuned
initial budget. Future work should evaluate live or replayed production traces,
different solver families, explicit latency objectives, and policies that
jointly choose initial and recovery budgets.

\section*{Ethical Considerations}

This work uses public mathematical reasoning benchmarks and introduces no new
user data or sensitive-domain dataset. The primary ethical concern is
reliability: post-generation reasoning can change correct answers into
incorrect ones. We therefore report harmful flips and recommend monitoring them
in deployed systems. Selective verification should not be treated as a
substitute for domain-specific validation in high-stakes applications.

\bibliography{main}

\appendix
\section*{Appendix}
\section{Intervention Screening}
\label{app:screening}

Before training the final selective gate, we screen three post-generation
actions on 2,000 MATH training examples. Each action observes the same base
attempt and receives a 4,096-token generation limit. Table~\ref{tab:screening}
reports the final answer accuracy and outcome transitions.

\begin{table*}[t]
\centering
\small
\setlength{\tabcolsep}{5pt}
\begin{tabular}{@{}lrrrrrr@{}}
\toprule
\textbf{Action} & \textbf{Examples} & \textbf{Final acc.} &
\textbf{Fix rate} & \textbf{Flip rate} & \textbf{Finalizer rate} &
\textbf{Avg. action tokens} \\
\midrule
\rowcolor{gray!7} Accept base & 2,000 & 78.85 & 0.00 & \costbest{0.00} & \costbest{0.00} & \costbest{0} \\
Continue & 2,000 & 89.75 & 13.25 & \risk{2.35} & 8.35 & 2,320 \\
Critique and repair & 2,000 & 89.80 & 12.40 & \risk{1.45} & 10.15 & 2,612 \\
\rowcolor{sevraGreen!7} Active verify & 2,000 & \best{91.35} & \best{13.70} & \risk{1.20} & 7.10 & 2,490 \\
\bottomrule
\end{tabular}
\caption{Full action screening on the MATH recovery-training set. Rates and
accuracy are percentages. Active verification provides the best static
accuracy and fix-to-flip trade-off.}
\label{tab:screening}
\end{table*}

The sampled-action oracle reaches 94.75\% accuracy, 3.40 points above the best
static action. This establishes that action selection has headroom, but it also
shows that the available headroom is smaller than the gain from simply invoking
active verification on every example. We choose active verification as the
primary action and train a binary gate to decide between accepting and
verifying. Independent solving was explored during pilot runs but was not
carried into the full screening because it produced substantially more harmful
answer changes and weaker accuracy.

\section{Exact Prompt Templates}
\label{app:prompts}

\promptbox{Base solve prompt}{
Solve the problem carefully and concisely. Do not discuss confidence. Finish
with exactly: \texttt{Final answer: <answer>}.\\
\texttt{Problem: <problem>}
}

\promptbox{Continue prompt}{
Continue from the attempted solution below. Check every operation and
assumption, then revise the answer if needed. Finish with exactly:
\texttt{Final answer: <answer>}.\\
\texttt{Problem: <problem>}\\
\texttt{Attempted solution: <base attempt>}
}

\promptbox{Critique-and-repair prompt}{
Audit the attempted solution. First identify the earliest concrete error, or
state that no concrete error is found. Then recompute the answer from that
point. Finish with exactly: \texttt{Final answer: <answer>}.\\
\texttt{Problem: <problem>}\\
\texttt{Attempted solution: <base attempt>}
}

\promptbox{Active-verification prompt}{
Create and execute at least two candidate-specific checks for the attempted
solution, such as reconstructing the governing equations, checking units or
bounds, substituting the result back, or solving by an independent route.
Preserve the answer if all checks pass; otherwise repair it. Finish with
exactly: \texttt{Final answer: <answer>}.\\
\texttt{Problem: <problem>}\\
\texttt{Attempted solution: <base attempt>}
}

\paragraph{Finalizer.}
If a reasoning-mode response exposes no final answer, we pass the tail of its
reasoning to a non-reasoning finalizer and request exactly one
\texttt{Final answer: <answer>} line. The finalizer receives at most 512
generation tokens. Its prompt and generation tokens are included in all
reported costs.

\section{Gate Inputs and Training}
\label{app:gates}

\subsection{Observable Features}
Feature definition are shown in Table \ref{tab:features}.
\begin{table}[ht]
\centering
\small
\resizebox{\columnwidth}{!}{
\begin{tabular}{p{0.28\columnwidth}p{0.62\columnwidth}}
\toprule
\textbf{Feature} & \textbf{Definition and availability} \\
\midrule
Completion reason & Whether the base request completed normally or reached its
generation limit; directly available from the inference server. \\
Generated tokens & Tokens generated by the base request; available from usage
metadata. \\
Finalizer use & Whether the base request required a separate answer-finalizer
call. \\
Difficulty & Static heuristic estimate attached to the example before solving. \\
Verification need & Static heuristic estimate of whether the task benefits
from checking. \\
Constraint density & Static estimate of the number and density of explicit
constraints in the problem. \\
Problem and attempt text & Used by the 0.6B and 1.7B sequence-classifier gates,
but not required by the cheap gate. \\
\bottomrule
\end{tabular}}
\caption{Inputs available to the recoverability gates. No feature uses a gold
answer or post-intervention outcome at deployment time.}
\label{tab:features}
\end{table}

\subsection{Learned Gate Objective}

The learned gate input asks the classifier to predict whether active
verification will correct the attempted solution. It includes task type,
estimated difficulty, verification need, constraint density, problem text, and
the base attempt. The binary label is the observed helpful-fix indicator.
Because helpful fixes are rare, training uses a positive-class weight equal to
the ratio of negative to positive training examples.

\subsection{QLoRA Hyperparameters}

\begin{table}[ht]
\centering
\small
\resizebox{\columnwidth}{!}{
\begin{tabular}{lr}
\toprule
\textbf{Setting} & \textbf{Value} \\
\midrule
Gate bases & Qwen3-0.6B, Qwen3-1.7B \\
Quantization & 4-bit NF4, double quantization \\
Compute dtype & FP16 \\
LoRA rank / alpha / dropout & 16 / 32 / 0.05 \\
Target modules & q, k, v, o, gate, up, down projections \\
Max input length & 1,536 tokens \\
Epochs & 3 \\
Learning rate & $2\times10^{-4}$ \\
Per-device batch size & 2 \\
Gradient accumulation & 8 \\
Development fraction & 20\%, split by example \\
Seed & 42 \\
Selection metric & Development AUPRC \\
\bottomrule
\end{tabular}
}
\caption{Learned-gate training configuration.}
\label{tab:gate-hparams}
\end{table}

The 0.6B gate reaches development AUROC 0.9567 and AUPRC 0.7601. The 1.7B
gate reaches AUROC 0.9570 and AUPRC 0.7534. Their best frozen thresholds are
0.09 and 0.023, respectively. Despite strong ranking metrics, both learned
gates remain close to the cheap feature gate in downstream policy quality.

\section{Data Construction and Evaluation Protocol}
\label{app:data}

\paragraph{Recovery training set.}
We use 2,000 MATH training examples, disjoint from \mathfive. Each example has
one shared base attempt and one rollout for accept, continue, critique-and-repair,
and active verification, giving 8,000 logged rows. Base accuracy is 78.85\%.
The base finalizer rate is 23.95\%. An example-level 80/20 split prevents
different actions from the same problem appearing in both gate training and
development sets.

\paragraph{Evaluation sets.}
\mathfive contains 500 examples. \gsm evaluation uses all 1,319 test examples.
No test labels are used to train gates or select thresholds. The \gsm
evaluation is frozen transfer: MATH-trained weights and thresholds are applied
without tuning.

\paragraph{Answer scoring.}
We first extract boxed or explicitly finalized answers. We then apply
normalization and numeric matching. For MATH-style expressions, we additionally
parse LaTeX and symbolic expressions and use mathematical-equivalence
verification. The same scorer is applied to base and intervention outputs.

\paragraph{Rollouts.}
Evaluation datasets contain two sampled intervention rollouts per action. The
policy analysis uses the logged intervention outcome consistently across
policies and performs paired resampling by example. The base attempt is shared
within each example, ensuring that comparisons isolate the action and routing
decision rather than base-generation noise.

\section{Full Budget and Cost Definitions}
\label{app:cost}

This appendix expands the compact cost notation used in the main paper. The
central distinction is between a \emph{configured} limit and a \emph{realized}
cost. Configured limits are maximum generation budgets supplied to the solver;
realized costs are the prompt-plus-generation tokens actually consumed by a
request, including answer-finalization calls when they occur. This distinction
is why a nominally larger long-base budget can be cheaper in practice: if the
model completes cleanly, it may avoid truncation, retries, and finalizers.

We also separate action tokens from total tokens. Action tokens measure only
the post-generation intervention, such as continuation or active verification.
Total tokens include the initial solve and any finalizer. This makes it harder
for a recovery method to look good by ignoring the base attempt it depends on.
For learned gates, we do not add classifier inference to language-model token
totals because the cheap gate has no solver-token cost and the QLoRA gates are
local classifiers. Their memory and latency costs remain real and are treated
as serving-stack considerations in Appendix~\ref{app:latency-protocol}.

\begin{table*}[t]
\centering
\small
\resizebox{\textwidth}{!}{
\begin{tabular}{lrrrr}
\toprule
\textbf{Policy component} & \textbf{Configured generation limit} &
\textbf{May use finalizer?} & \textbf{Included in action tokens?} &
\textbf{Included in total tokens?} \\
\midrule
Short base & 4,096 & Yes, 512-token limit & No & Yes \\
Long base & 8,192 & Yes, 512-token limit & No & Yes \\
Continue & 4,096 & Yes, 512-token limit & Yes & Yes \\
Critique and repair & 4,096 & Yes, 512-token limit & Yes & Yes \\
Active verification & 4,096 & Yes, 512-token limit & Yes & Yes \\
Gate inference & N/A & No & No & No \\
\bottomrule
\end{tabular}
}
\caption{Budget and accounting definitions. Gate inference cost is excluded
from model-token totals because the cheap gate has no language-model tokens and
the learned gates run locally as classifiers. Their memory and latency costs
remain an operational consideration.}
\label{tab:cost-def}
\end{table*}

\section{Full Policy Results}
\label{app:full-results}

The main table compresses the headline policies into one view; this appendix
keeps the full policy rows with confidence intervals and action-token columns.
The rows should be read as a matched serving comparison. Accepting the base
answer consumes no action tokens. Always-on policies spend an extra call on
every example. Selective policies spend that call only when the gate score
crosses the frozen threshold. Long-base policies spend no post-generation call,
but change the initial maximum budget.

The most important pattern is that action-level efficiency and system-level
efficiency can disagree. Selective verification is clearly better than always
verifying as a post-generation intervention, because it preserves most fixes
while removing many unnecessary calls. However, total-token accounting shows
that a longer initial solve can still dominate the overall cost frontier. This
is the paper's main deployment warning: recovery should be evaluated against a
tuned initial-budget baseline, not only against an always-recover baseline.

\begin{table*}[t]
\centering
\small
\begin{tabular}{lrrrrrr}
\toprule
\textbf{\mathfive policy} & \textbf{Acc.} & \textbf{95\% CI} &
\textbf{Intervene} & \textbf{Action tok.} & \textbf{Total tok.} &
\textbf{Flips} \\
\midrule
Base, 4,096 limit & 59.0 & [54.8, 63.2] & 0.0 & 0 & 4,313 & 0.0 \\
Always continue & 72.0 & [68.4, 75.6] & 100.0 & 3,694 & 8,007 & 3.6 \\
Selective continue & 73.6 & [69.9, 77.2] & 48.2 & 2,751 & 7,064 & 1.7 \\
Always active verify & 75.5 & [71.9, 78.8] & 100.0 & 3,812 & 8,125 & 2.2 \\
Selective active verify & 76.3 & [72.8, 79.8] & 48.2 & 2,791 & 7,104 & 1.0 \\
Long base, 8,192 limit & 76.0 & [72.2, 79.6] & 0.0 & 0 & 5,124 & 0.0 \\
\bottomrule
\end{tabular}
\caption{Complete \mathfive policy results. Accuracy, intervention, and flip
rates are percentages.}
\label{tab:math-full}
\end{table*}

\begin{table*}[t]
\centering
\small
\begin{tabular}{lrrrrr}
\toprule
\textbf{\gsm policy} & \textbf{Acc.} & \textbf{Intervene} &
\textbf{Action tok.} & \textbf{Total tok.} & \textbf{Flips} \\
\midrule
Base, 4,096 limit & 93.40 & 0.0 & 0 & 1,180 & 0.00 \\
Always active verify & 93.40 & 100.0 & 1,752 & 2,932 & 1.25 \\
Selective active verify & 94.47 & 3.0 & 154 & 1,335 & 0.00 \\
Long base, 8,192 limit & 94.54 & 0.0 & 0 & 1,157 & 0.00 \\
\bottomrule
\end{tabular}
\caption{Complete \gsm transfer results for the primary 1.7B selective gate.}
\label{tab:gsm-full}
\end{table*}

The \gsm table is especially useful as a transfer diagnostic. The selective
gate is trained and thresholded on MATH development data, then applied to GSM8K
without recalibration. The gate verifies only a small subset, which is exactly
what a production controller should do when base accuracy is already high.
Always verifying in this setting is not just wasteful; it introduces observed
right-to-wrong flips that the selective policy avoids.

\section{Paired Comparisons}
\label{app:statistics}

Aggregate accuracy alone hides whether a policy is repairing failures or
damaging successes. The paired comparisons below therefore report accuracy,
flip, and token differences on the same examples. Confidence intervals use
paired bootstrap resampling by example so that the correlation between two
policies is preserved. For token-reduction rows, the interval is reported as a
relative action-token reduction because that is the operational quantity a
serving team would use to estimate intervention load.

\begin{table*}[t]
\centering
\small
\resizebox{\textwidth}{!}{
\begin{tabular}{llrrr}
\toprule
\textbf{Dataset} & \textbf{Comparison} & \textbf{Difference} &
\textbf{95\% paired-bootstrap CI} & \textbf{$p$} \\
\midrule
\mathfive & Selective verify minus always verify accuracy &
+0.8 points & [$-0.1$, +1.7] & .103 \\
\mathfive & Selective verify minus always verify flips &
$-1.2$ points & [$-1.6$, $-0.4$] & $<.001$ \\
\mathfive & Selective verify action-token reduction vs.\ always verify &
26.8\% & [23.0\%, 30.8\%] & -- \\
\gsm & Selective verify minus always verify accuracy &
+1.06 points & [+0.53, +1.63] & $<.001$ \\
\gsm & Selective verify minus always verify flips &
$-1.25$ points & [$-1.78$, $-0.80$] & $<.001$ \\
\gsm & Selective verify action-token reduction vs.\ always verify &
91.2\% & [88.6\%, 93.9\%] & -- \\
\gsm & Selective verify minus long-base accuracy &
$-0.08$ points & [$-0.91$, +0.80] & .899 \\
\gsm & Selective verify minus long-base total tokens &
+178 tokens & [+105, +256] & $<.001$ \\
\gsm & Long base minus short base accuracy &
+1.14 points & [+0.15, +2.12] & .022 \\
\bottomrule
\end{tabular}
}
\caption{Key paired comparisons. Differences use selective policy minus the
named comparator unless otherwise stated.}
\label{tab:paired}
\end{table*}

Two conclusions follow from these paired tests. First, selective verification's
main statistically robust win over always verifying is reliability and cost:
it significantly reduces harmful flips and action tokens on both math
benchmarks. Second, the long-base comparison is not a weak baseline. On GSM8K,
selective verification and long-base accuracy are statistically
indistinguishable, while long-base uses fewer total tokens. This is why the
paper frames \sevra as a recovery controller for settings where explicit
verification, retry control, or answer-change auditing matters. In other
settings, a tuned initial budget may be the simpler serving choice.

\section{Attempt-State Subgroups}
\label{app:subgroups}

\begin{table*}[t]
\centering
\small
\begin{tabular}{llrrrrr}
\toprule
\textbf{Dataset} & \textbf{Attempt state} & \textbf{$n$} & \textbf{Base} &
\textbf{Always verify} & \textbf{Selective verify} & \textbf{Always continue} \\
\midrule
\multirow{2}{*}{\mathfive}
& Completed & 273 & 98.17 & 97.80 & \textbf{98.90} & 96.15 \\
& Truncated & 227 & 11.89 & 48.68 & \textbf{49.12} & 42.95 \\
\midrule
\multirow{2}{*}{\gsm}
& Completed & 1,281 & \textbf{95.71} & 94.61 & \textbf{95.71} & -- \\
& Truncated & 38 & 15.79 & \textbf{52.63} & \textbf{52.63} & -- \\
\bottomrule
\end{tabular}
\caption{Accuracy by observable base-attempt completion state. The selective
policy preserves completed attempts and concentrates recovery on truncated
attempts.}
\label{tab:subgroups}
\end{table*}

For \mathfive, 45.4\% of base attempts are truncated, making recovery a common
need. For \gsm, only 2.9\% are truncated. The frozen MATH gate transfers because
it largely recognizes this observable failure mode. This is useful but also
limits the breadth of the claim: the results do not establish a general
semantic-error detector.

\section{Gate Baselines and Oracle Headroom}
\label{app:gate-baselines}

\begin{table*}[t]
\centering
\small
\begin{tabular}{lrrrr}
\toprule
\textbf{\mathfive routing rule} & \textbf{Acc.} & \textbf{Verify rate} &
\textbf{Action tokens} & \textbf{Flips} \\
\midrule
Always verify & 75.5 & 100.0 & 3,812 & 2.2 \\
Frozen cheap feature gate & 75.9 & 45.0 & 2,719 & 1.0 \\
Completion-risk proxy & 74.9 & 64.0 & 3,223 & 2.0 \\
Verification-need heuristic & 74.9 & 91.4 & 3,664 & 2.2 \\
Difficulty heuristic & 74.9 & 91.4 & 3,664 & 2.2 \\
Qwen3-0.6B gate & 75.7 & 46.4 & 2,760 & 1.2 \\
Qwen3-1.7B gate & \textbf{76.3} & 48.2 & 2,791 & 1.0 \\
Matched-rate value oracle & 77.7 & 48.2 & 1,807 & 0.0 \\
\bottomrule
\end{tabular}
\caption{Gate and heuristic baselines on \mathfive. The oracle indicates
remaining headroom if intervention value could be predicted perfectly at the
same verification rate.}
\label{tab:gate-baselines}
\end{table*}

The matched-rate oracle leads the primary gate by 1.4 accuracy points while
using fewer action tokens. Better value estimation could therefore improve
selection, but the larger opportunity in our experiments comes from initial
budget allocation.

\section{Failure Taxonomy}
\label{app:failures}

Manual inspection of logged attempts suggests four recurring outcome types:
\begin{enumerate}
    \item \textbf{Truncated derivation}: the base request reaches its token
    limit before exposing a final answer. Verification or a longer initial
    budget often recovers it.
    \item \textbf{Local arithmetic or algebra error}: the base attempt completes
    but contains a checkable operation error. Candidate-specific verification
    can repair it.
    \item \textbf{Incorrect reinterpretation}: a post-generation call replaces a
    correct answer after inventing an error or changing the problem
    interpretation. This produces a harmful flip.
    \item \textbf{Shared misconception}: the base and verification calls agree
    on the same wrong derivation. Additional calls do not help because the
    failure is correlated.
\end{enumerate}
Active verification is particularly useful for the first two categories.
Selective routing reduces exposure to the third, but does not solve the fourth.

\begin{table*}[t]
\centering
\small
\resizebox{\textwidth}{!}{
\begin{tabular}{p{0.18\textwidth}p{0.31\textwidth}p{0.20\textwidth}p{0.22\textwidth}}
\toprule
\textbf{Logged case} & \textbf{Problem sketch} & \textbf{Observed transition} & \textbf{Takeaway} \\
\midrule
\mathfive \#6 & Smallest positive perfect cube expressible as the sum of three consecutive integers. & Base $3$ wrong; active verification returns $27$ correct. & Candidate-specific checks can repair a local mathematical miss. \\
\mathfive \#4 & Cross-country graph asks which student has the greatest average speed. & Base ``Evelyn'' correct; continuation changes to numeric ``3.6'' wrong. & Extra reasoning can answer a different latent question. \\
\gsm \#7 & Interrupted 200GB download with restart overhead. & Base $20$ wrong; active verification returns $160$ correct. & Verification helps when the base attempt misses a compositional constraint. \\
CommonsenseQA \#12 & Multiple-choice question asking where a heifer's master lives. & Base option A correct; active verification flips to option E wrong. & Verification prompts can overthink short commonsense answers. \\
\bottomrule
\end{tabular}}
\caption{Representative logged outcomes. The examples illustrate why the
controller is framed around recoverability and harmful flips rather than
assuming that every additional reasoning call is beneficial.}
\label{tab:qualitative}
\end{table*}

\section{Negative Results and Design Decisions}
\label{app:negative}

\paragraph{Continuation is not a cheaper substitute for verification.}
Selective continuation and selective verification consume nearly identical
total tokens on \mathfive, yet verification is 2.7 points more accurate and
causes fewer flips. This rejects the hypothesis that any additional reasoning
call is sufficient.

\paragraph{A larger gate is not clearly better.}
The 1.7B gate obtains the best \mathfive result, but its advantage over the
cheap feature gate is only 0.4 points and disappears on \gsm. Model-gate
capacity is not the main bottleneck in the tested setting.

\paragraph{Always verifying does not reliably improve accuracy.}
On \gsm, always verifying makes enough harmful changes to cancel its helpful
fixes. This rejects the assumption that verification is harmless.

\paragraph{Selective verification is not the most efficient overall policy.}
The long initial solve matches it with fewer total tokens on both benchmarks.
The appropriate claim is that selective verification is the strongest tested
\emph{post-generation} intervention and a useful recovery mechanism, not the
universal cost winner.

\section{Reproducibility and Artifact Map}
\label{app:repro}

\begin{table*}[t]
\centering
\small
\begin{tabular}{p{0.27\textwidth}p{0.65\textwidth}}
\toprule
\textbf{Artifact type} & \textbf{Contents} \\
\midrule
Recovery generation & Scripts for base attempts, interventions, usage logging,
answer extraction, finalizers, and resumable JSONL output. \\
Data integrity & Shard merge and deduplication scripts; summaries warn on
partial examples and exclude incomplete action sets. \\
Gate training & QLoRA sequence-classifier training for 0.6B and 1.7B gates,
including example-level splits and saved policy curves. \\
Gate evaluation & Prediction export, frozen-threshold evaluation, heuristic
baselines, matched-rate comparisons, and oracle analysis. \\
Decisive baselines & Scripts and configs for continuation, active verification,
short-base, and 8,192-token long-base runs. \\
Statistical analysis & Paired bootstrap confidence intervals, significance
tests, token reductions, harmful flips, and completed/truncated subgroups. \\
Paper figures & Self-contained TikZ sources using values from the final JSON
summaries. \\
Public replay & Static Hugging Face Space at
\url{https://huggingface.co/spaces/sevra-space/sevra-replay}; includes
precomputed tables, cost--accuracy visualization, fix/flip visualization, and
representative logged examples. \\
\bottomrule
\end{tabular}
\caption{Reproducibility inventory. Anonymous artifact paths can be released
with the paper artifact.}
\label{tab:artifact-map}
\end{table*}

\paragraph{Execution environment.}
The final experiments use Python 3.10.12, Ollama 0.23.0, and two NVIDIA TITAN
RTX GPUs with 24GB memory each. Dual-GPU generation runs independent shards and
then merges rows by example and action identifiers. Runs are resumable: already
completed rows are skipped.

\paragraph{Seeds and thresholds.}
All reported experiments use seed 42. Gate thresholds are selected on the MATH
development split and frozen before test evaluation. The \gsm transfer uses the
same checkpoints and thresholds.

\section{Public Replay Dashboard}
\label{app:public-replay}

To make the paper's serving-style claims easier to audit, we deployed a public
static replay dashboard:
\url{https://huggingface.co/spaces/sevra-space/sevra-replay}. The direct static
application is available at
\url{https://sevra-space-sevra-replay.static.hf.space/index.html}. The
dashboard is intentionally static: it does not run Qwen3-4B, does not contact
Ollama, does not require GPU resources, and does not expose private logs. It
only visualizes precomputed aggregate results and a small set of representative
examples that are already described qualitatively in the paper.

\paragraph{Why static deployment?}
The evaluated solver stack uses long-context local generation and two TITAN RTX
GPUs. A free hosted CPU service is not an appropriate environment for faithfully
serving that solver. A static replay is therefore more stable than a slow or
inconsistent live demo: it makes the cost--accuracy and fix--flip
trade-offs inspectable while avoiding misleading claims about production
latency or hosted model availability. The artifact should be read as
\emph{deployment-adjacent evidence}: it demonstrates how the results would be
presented in a serving-style dashboard, not that the controller is deployed in
a live product.

\begin{table*}[t]
\centering
\small
\begin{tabular}{p{0.22\textwidth}p{0.67\textwidth}}
\toprule
\textbf{Dashboard panel} & \textbf{Purpose} \\
\midrule
Deployment takeaway & Summarizes the operational conclusion: tune the initial
budget first, then use selective recovery when explicit checks or regression
risk control matter. \\
Cost--accuracy frontier & Shows average realized model tokens and accuracy for
short base, long base, always-on intervention, selective intervention, and
Self-Consistency@5 where applicable. \\
Main tables & Reproduces the paper's core numbers for \mathfive, \gsm, and
CommonsenseQA. \\
Helpful fixes vs. harmful flips & Visualizes wrong-to-right and right-to-wrong
answer changes, emphasizing that extra reasoning has asymmetric reliability
risk. \\
Logged example patterns & Gives compact qualitative examples of useful
verification, harmful continuation, GSM recovery, and CommonsenseQA
overthinking. \\
\bottomrule
\end{tabular}
\caption{Contents of the deployed static replay dashboard. The dashboard is
designed for transparent inspection of aggregate serving metrics rather than live
model interaction.}
\label{tab:dashboard-contents}
\end{table*}

\paragraph{Anonymity and artifact use.}
The Space is hosted from an anonymous project account and contains no author
names, institutional identifiers, machine paths, or hidden run directories. If
an anonymous URL is not allowed by the venue policy, the same files
can be submitted as supplementary material: the dashboard is a single
\texttt{index.html} file plus metadata.

\section{Formal Controller Logic}
\label{app:formal-controller}

This section spells out the serving decision implemented by \sevra. Let
$x_i$ denote an input problem, $y_i$ its gold answer, and $b_i$ the base solver
attempt. Let $a_i^{0}$ be the answer extracted from the base attempt, and let
$m_i$ be the observable metadata for that attempt: realized tokens, configured
budget, completion status, finalizer status, and lightweight text statistics.
A gate $g_{\theta}$ maps metadata and visible attempt text to a recoverability
score,
\begin{equation}
    s_i = g_{\theta}(x_i, b_i, m_i) \in [0,1].
\end{equation}
Given a threshold $\tau$, the selective policy is
\begin{equation}
    \pi_{\tau}(x_i) =
    \begin{cases}
        \textsc{active\_verify}, & s_i \ge \tau,\\
        \textsc{accept}, & s_i < \tau.
    \end{cases}
\end{equation}
If the policy accepts, the final answer is $a_i^{0}$. If it verifies, the
system runs an active-verification prompt and extracts a new answer
$a_i^{v}$. The deployed final answer is therefore
\begin{equation}
    \hat{y}_i(\pi_{\tau}) =
    \mathbb{I}[s_i < \tau] a_i^{0}
    + \mathbb{I}[s_i \ge \tau] a_i^{v},
\end{equation}
where the indicator notation denotes a branch choice rather than numerical
addition of strings.

\paragraph{Recoverability labels.}
For a candidate action $k$, define correctness
\begin{equation}
    c(a_i^k, y_i) =
    \mathbb{I}[\operatorname{match}(a_i^k, y_i)].
\end{equation}
The positive label for a recovery action is not simply ``the action is
correct.'' Instead, it captures whether the action repairs the base attempt:
\begin{equation}
    z_i^k =
    \mathbb{I}\left[c(a_i^0,y_i)=0 \wedge c(a_i^k,y_i)=1\right].
\end{equation}
This label targets operational recovery. It asks whether an extra call is
useful relative to accepting the already-computed answer. A separate harmful
flip indicator captures the opposite risk:
\begin{equation}
    h_i^k =
    \mathbb{I}\left[c(a_i^0,y_i)=1 \wedge c(a_i^k,y_i)=0\right].
\end{equation}

\paragraph{Threshold choice.}
Thresholds are selected on the held-out MATH development split by maximizing
downstream policy accuracy under the logged action outcomes. Ties are broken
by lower realized action tokens and then lower harmful-flip rate. This differs
from selecting a classifier threshold by F1: the serving decision is asymmetric,
because false positives consume an additional model call and can damage a
correct answer.

\begin{table*}[t]
\centering
\small
\begin{tabular}{p{0.24\textwidth}p{0.36\textwidth}p{0.30\textwidth}}
\toprule
\textbf{Object} & \textbf{Definition} & \textbf{Why it matters} \\
\midrule
$a_i^0$ & Answer extracted from the base attempt & The zero-extra-call
deployment default. \\
$a_i^v$ & Answer extracted after active verification & Candidate output if
the controller intervenes. \\
$s_i$ & Gate score from visible attempt state and metadata & Determines whether
the system spends another call. \\
$z_i^v$ & Wrong-to-right repair label & Trains the gate toward recoverable
failures rather than generic uncertainty. \\
$h_i^v$ & Right-to-wrong harmful flip & Measures regression risk from extra
reasoning. \\
$\tau$ & Operating threshold & Converts scores into a concrete serving policy. \\
\bottomrule
\end{tabular}
\caption{Notation used in the formal controller definition.}
\label{tab:controller-notation}
\end{table*}

\section{Metric Definitions}
\label{app:metrics}

The paper reports aggregate accuracy, intervention rate, token cost, helpful
fixes, harmful flips, and oracle headroom. All quantities are computed at the
example level before averaging, so paired bootstrap resampling can preserve the
correlation between policies.

\paragraph{Accuracy.}
For a policy $\pi$, accuracy over $N$ examples is
\begin{equation}
    \operatorname{Acc}(\pi) =
    \frac{1}{N}\sum_{i=1}^{N}
    c(\hat{y}_i(\pi),y_i).
\end{equation}
For MATH and GSM8K, $c$ combines exact answer normalization with mathematical
equivalence checking. For CommonsenseQA, $c$ checks the final multiple-choice
label.

\paragraph{Intervention rate.}
The intervention rate is the fraction of examples on which the controller runs
an additional action:
\begin{equation}
    \operatorname{IR}(\pi) =
    \frac{1}{N}\sum_{i=1}^{N}
    \mathbb{I}[\pi(x_i)\ne \textsc{accept}].
\end{equation}
This rate is important because two policies with similar accuracy may create
very different operational loads.

\paragraph{Helpful fixes and harmful flips.}
For policy $\pi$, helpful fixes and harmful flips are
\begin{align}
\operatorname{Fix}(\pi)
={}& \frac{1}{N}\sum_{i=1}^{N}
\mathbb{I}\Big[
c(a_i^0,y_i)=0 \\
&\qquad\qquad \wedge\
c(\hat{y}_i(\pi),y_i)=1
\Big], \notag\\
\operatorname{Flip}(\pi)
={}& \frac{1}{N}\sum_{i=1}^{N}
\mathbb{I}\Big[
c(a_i^0,y_i)=1 \\
&\qquad\qquad \wedge\
c(\hat{y}_i(\pi),y_i)=0
\Big].
\end{align}
The net accuracy change relative to accepting the base answer can be written as
\begin{equation}
    \operatorname{Acc}(\pi) - \operatorname{Acc}(\textsc{accept})
    = \operatorname{Fix}(\pi) - \operatorname{Flip}(\pi).
\end{equation}
This identity is one reason we emphasize flips: an intervention that produces
many fixes can still be unattractive if it also creates many regressions.

\paragraph{Token accounting.}
Let $T_i^0$ be realized prompt-plus-generation tokens for the base attempt,
$T_i^k$ the realized tokens for action $k$, and $F_i$ the realized finalizer
tokens if a finalizer is invoked. Total model tokens for a policy are
\begin{equation}
\begin{aligned}
\operatorname{Tok}(\pi)
={}& \frac{1}{N}\sum_{i=1}^{N}
\Bigg(
T_i^0 + F_i^0 \\
&\quad + \sum_{k \in \mathcal{A}}
\mathbb{I}[\pi(x_i)=k]
\left(T_i^k+F_i^k\right)
\Bigg).
\end{aligned}
\end{equation}
This quantity differs from configured maximum budget. A run with an 8,192-token
limit can be cheaper than a 4,096-token run plus verification if it usually
finishes early and avoids finalizers.

\paragraph{Oracle headroom.}
The sampled-rollout oracle selects the best observed action for each example:
\begin{equation}
    \operatorname{Oracle}_{\text{sample}}(i)
    = \max_{k \in \mathcal{A}_{i}}
    c(a_i^k,y_i),
\end{equation}
where $\mathcal{A}_{i}$ contains the base answer and logged intervention
answers for example $i$. The expected-action oracle first estimates expected
correctness for each action family from the sampled rows, then selects the
best action family per example. Both oracles use gold labels and are therefore
diagnostics, not deployable policies.

\paragraph{Efficiency summaries.}
Some internal summaries use confidence-weighted or success-weighted
accuracy-per-token scores. The paper does not rely on these as primary
claims, because they compress reliability and cost into one number. We report
the underlying accuracy, token, fix, and flip quantities instead.

\section{Logged Row Schema}
\label{app:logging-schema}

Each recovery dataset row corresponds to one example-action pair. Accept rows
store the base attempt. Intervention rows store the additional action result
while sharing the same base attempt identifier. This layout enables off-policy
evaluation of many controllers without rerunning the expensive solver.

\begin{table*}[t]
\centering
\small
\begin{tabular}{p{0.24\textwidth}p{0.18\textwidth}p{0.46\textwidth}}
\toprule
\textbf{Field} & \textbf{Type} & \textbf{Description} \\
\midrule
\texttt{example\_id} & integer/string & Stable identifier used for merging
action rows and paired bootstrap resampling. \\
\texttt{dataset} & string & Dataset name such as \texttt{math500},
\texttt{gsm8k}, or \texttt{commonsenseqa}. \\
\texttt{question} & string & Input prompt after benchmark-specific
formatting. \\
\texttt{gold} & string & Normalized reference answer or multiple-choice
label. \\
\texttt{action} & string & One of accept, continue, critique-repair,
active-verify, long-base, or self-consistency sample. \\
\texttt{base\_answer} & string & Extracted answer from the shared base
attempt. \\
\texttt{action\_answer} & string & Extracted answer after the intervention;
empty for accept rows. \\
\texttt{final\_answer} & string & Answer used by the evaluated policy row. \\
\texttt{base\_correct} & boolean & Correctness of the base answer. \\
\texttt{final\_correct} & boolean & Correctness after selecting the row's
action. \\
\texttt{finish\_reason} & string & Completion state exposed by the model
runtime, including length-limit termination when available. \\
\texttt{finalizer\_used} & boolean & Whether an answer-extraction finalizer was
called because the reasoning output lacked a parseable final response. \\
\texttt{prompt\_tokens} & integer & Realized input tokens for the relevant
model call. \\
\texttt{completion\_tokens} & integer & Realized generated tokens for the
relevant model call. \\
\texttt{total\_tokens} & integer & Prompt plus generated tokens, including
finalizer calls when applicable. \\
\texttt{trace\_path} & string & Relative path to the saved trace for audit and
debugging. \\
\bottomrule
\end{tabular}
\caption{Canonical logged-row schema used by the recovery and evaluation
pipeline. Individual scripts may store additional debug fields, but these
fields are sufficient to reproduce the policy comparisons in the paper.}
\label{tab:logging-schema}
\end{table*}

\paragraph{Partial examples.}
Generation jobs are sharded and resumable. A partially completed example can
occur if one action finishes but another action is interrupted. Summary scripts
therefore check the expected number of rows per example and exclude incomplete
sets from oracle or static-action summaries. The console warnings reported
during the experiments are intentional safeguards rather than silent failures.

\paragraph{Deduplication.}
Shard merge scripts deduplicate by example identifier, action name, and sample
index. If duplicate rows exist, the first completed row is preserved and later
duplicates are ignored. This conservative rule avoids allowing retries to
silently improve a policy after the fact.

\section{Evaluation Flow}
\label{app:evaluation-flow}

The complete evaluation path has seven stages. First, the base solver produces
a reasoning attempt under a configured maximum budget. Second, answer
extraction is attempted directly from the response. Third, if direct extraction
fails, a non-reasoning finalizer is called to produce a parseable final answer.
Fourth, each candidate intervention is run from the same base attempt. Fifth,
base and intervention answers are normalized and scored. Sixth, gate scores are
joined onto the logged rows. Seventh, each policy is evaluated by selecting the
appropriate row for every example and then computing paired metrics.

\begin{table*}[t]
\centering
\small
\begin{tabular}{p{0.12\textwidth}p{0.27\textwidth}p{0.48\textwidth}}
\toprule
\textbf{Stage} & \textbf{Output} & \textbf{Failure mode checked} \\
\midrule
Base solve & Base trace, answer, usage & Solver truncation, missing final
answer, unexpectedly empty output. \\
Finalization & Parseable base answer & Finalizer overuse, finalizer changing a
completed answer, answer-format mismatch. \\
Intervention & Action trace, answer, usage & Extra reasoning produces no
answer, repeats base mistake, or changes a correct answer. \\
Scoring & Correctness flags & Normalization mismatch, symbolic-equivalence
failure, multiple-choice label mismatch. \\
Gate export & Recoverability scores & Missing features, train/test leakage,
threshold accidentally tuned on test labels. \\
Policy replay & Final policy answer & Incorrect row selection, duplicate rows,
partial examples. \\
Summary & Tables, figures, CIs & Aggregate-only claims hiding flips or cost
shifts. \\
\bottomrule
\end{tabular}
\caption{Evaluation stages and the main failure mode checked at each stage.}
\label{tab:evaluation-flow}
\end{table*}

\paragraph{Why finalizers are counted.}
Some long reasoning outputs contain a correct derivation but omit a clean final
answer marker, while some truncated outputs end mid-calculation. A finalizer is
therefore part of the serving path, not an external evaluator. Counting its
tokens prevents a policy from looking artificially cheap by producing
unparseable responses that require a second model call.

\paragraph{Why base and intervention share an example key.}
The policy question is counterfactual: for the same base attempt, should the
system accept, continue, critique-repair, or actively verify? Sharing the base
attempt makes these choices comparable. Without this shared key, differences
between actions would be confounded by stochastic variation in the first solve.

\section{Controller Implementation Details}
\label{app:implementation-details}

The implementation separates action generation from policy evaluation. Action
generation is expensive and model-facing; policy evaluation is cheap and
offline. This separation is important for fast iteration: a new gate threshold
or cheap-feature baseline can be evaluated without rerunning Qwen3-4B.

\paragraph{Cheap gate.}
The cheap gate uses only serving-visible information: completion state, token
usage, finalizer use, answer length, answer format, and shallow prompt/response
statistics. It is intended as the minimum operational baseline for any learned
router. If a large learned gate does not beat this baseline by a meaningful
margin, the learned gate is unlikely to justify its serving complexity.

\paragraph{Learned gates.}
The 0.6B and 1.7B gates are trained as binary sequence classifiers with QLoRA.
Inputs combine the question, the base answer, compact base-attempt metadata,
and visible base-attempt text. The target is the active-verification repair
label $z_i^v$. Because fixes are rarer than accepts, checkpoint selection uses
development AUPRC before downstream policy thresholding.

\paragraph{Action prompts.}
The active-verification prompt is candidate-specific: it asks the model to
check the base answer against the problem and either preserve or repair it. The
continuation prompt extends the existing attempt. The critique-repair prompt
asks for an error analysis followed by a corrected answer. These actions are
deliberately simple, because the paper isolates whether selective recovery is
worthwhile rather than optimizing prompt engineering for every benchmark.

\paragraph{Selection invariants.}
The replay evaluator enforces four invariants:
\begin{enumerate}
    \item every evaluated policy selects at most one final answer per example;
    \item accept policies never consume action tokens;
    \item selective policies consume action tokens only above threshold;
    \item all paired comparisons use the same example set.
\end{enumerate}
These invariants keep accuracy, token, and flip comparisons aligned.

\section{Reproduction Commands and File Map}
\label{app:commands}

The exact repository layout may differ in an anonymous release, but the
experiments are organized around four command families:
\begin{enumerate}
    \item generate logged recovery rows for each action and shard;
    \item merge shards and remove incomplete example-action sets;
    \item train/export gate scores and thresholded policies;
    \item summarize tables, figures, paired tests, and the replay dashboard.
\end{enumerate}

\begin{table*}[t]
\centering
\small
\begin{tabular}{p{0.23\textwidth}p{0.25\textwidth}p{0.40\textwidth}}
\toprule
\textbf{Component} & \textbf{Representative path} & \textbf{Role} \\
\midrule
Generation scripts & \texttt{scripts/} & Launch base, continuation,
critique-repair, active-verification, long-base, and self-consistency runs. \\
Experiment configs & \texttt{configs/} & Store dataset paths, model names,
budget limits, gate thresholds, and output directories. \\
Runtime code & \texttt{src/} & Executor wrappers, answer extraction,
normalization, usage accounting, and policy replay utilities. \\
Oracle analysis & \texttt{ablation\_oracle.py} & Computes sampled-rollout and
expected-action oracle summaries from logged rows. \\
Recovery outputs & \texttt{outputs/recovery/} & JSONL rows for base and
intervention outcomes. \\
Result summaries & \texttt{outputs/results/} & Aggregate JSON/CSV summaries,
confidence intervals, and paper-ready numbers. \\
Trace outputs & \texttt{outputs/traces/} & Per-example model traces for audit
and debugging. \\
Replay dashboard & \texttt{sevra\_hf\_space/} & Static dashboard source used by
the public Hugging Face Space. \\
\bottomrule
\end{tabular}
\caption{Implementation map for reproducing the experiments and deployed
replay dashboard.}
\label{tab:file-map}
\end{table*}

\paragraph{Minimal reproducibility recipe.}
A minimal independent reproduction does not require retraining every gate. It
can start from released recovery JSONL files, verify row completeness, compute
base/always/selective/long-base policies, and regenerate the paper tables.
Full reproduction additionally reruns model generation and QLoRA gate training.
The distinction matters because generation is the slowest and most expensive
part of the pipeline, while policy replay is fast.

\paragraph{Expected runtime bottlenecks.}
The slowest stage is not logistic routing or summary generation; it is
long-context local model inference through the solver API. Sharding across two
GPUs gives near-linear throughput only when each shard keeps the model loaded
and avoids frequent process restarts. Self-Consistency@5 is especially
expensive because it multiplies calls even when individual answers are short.

\section{Additional Validity Checks}
\label{app:validity-checks}

\paragraph{No test-label threshold tuning.}
Thresholds are chosen on the held-out MATH development split. The \mathfive
test set, \gsm test set, and CommonsenseQA validation set are used only for
final evaluation. The \gsm result is intentionally a frozen transfer test: no
\gsm labels are used to recalibrate the gate.

\paragraph{Long-base comparison.}
The 8,192-token baseline is included because any controller that adds a second
call should be compared against spending more budget up front. This baseline is
especially important for Industry Track claims: a controller that saves
verification tokens may still be dominated by a simpler serving configuration.

\paragraph{Workload transfer.}
CommonsenseQA is included to test whether the verification story transfers
outside math. It does not. Active verification lowers accuracy and increases
harmful flips, while Self-Consistency@5 improves accuracy at much higher cost.
This supports a workload-specific deployment message rather than a universal
verification message.

\paragraph{Dashboard consistency check.}
The static replay dashboard uses the same aggregate numbers as the final paper
tables. It is intentionally not connected to a live model, so it cannot drift
because of package versions, model updates, or server availability. Readers can
inspect the trade-off surface immediately, while full run reproduction can be
done separately from the released code and JSONL files.

\section{Initial-Budget Frontier}
\label{app:budget-frontier}

A single 8,192-token baseline is not enough to characterize the initial-budget
frontier. We therefore include an intermediate 6,144-token MATH500 baseline.
The trend is monotonic: larger
initial budgets reduce truncation and improve accuracy, while realized token
growth is much smaller than the configured maximum-token growth because fewer
examples need a finalizer.

\begin{table*}[h]
\centering
\small
\begin{tabular}{lrrrr}
\toprule
\textbf{Initial budget} & \textbf{Acc.} & \textbf{Avg. total tokens} &
\textbf{Finalizer rate} & \textbf{Truncation rate} \\
\midrule
4,096 & 59.0 & 4,313 & 45.2 & 45.4 \\
6,144 & 68.0 & 4,759 & 30.2 & 32.0 \\
8,192 & 76.0 & 5,124 & 21.6 & 22.8 \\
\bottomrule
\end{tabular}
\caption{\mathfive initial-budget sweep. The 6,144-token point confirms that
the long-base result is not an isolated artifact: increasing initial budget
smoothly improves accuracy and reduces truncation.}
\label{tab:budget-frontier}
\end{table*}

Paired bootstrap differences are significant for adjacent budget points:
6,144 improves over 4,096 by 9.0 points (95\% CI [5.2, 12.8]) at an average
cost of 447 additional realized tokens, while 8,192 improves over 6,144 by 8.0
points (95\% CI [4.8, 11.2]) at an average cost of 365 additional realized
tokens. This strengthens the paper's main caution: an adaptive recovery method
should be compared against a tuned initial budget, not only against a short
fixed-budget baseline.

\section{Operational Translation for Deployment}
\label{app:operational}

Table~\ref{tab:operational} translates the main comparisons into quantities a
serving team would monitor per 1,000 requests. These numbers are not provider
prices or latency measurements; they are derived directly from the realized
model-token accounting in Tables~\ref{tab:math500}--\ref{tab:gsm8k}. The
purpose is to make the deployment trade-offs auditable: how many extra calls
are avoided, how many harmful answer changes are prevented, and how much model
work moves between the initial solve and post-generation recovery.

\begin{table*}[h]
\centering
\small
\resizebox{\textwidth}{!}{
\begin{tabular}{p{0.25\textwidth}p{0.66\textwidth}}
\toprule
\textbf{Comparison} & \textbf{Operational interpretation per 1,000 requests} \\
\midrule
\mathfive selective verification vs. always verify & Avoids 518 verification calls, saves about 1.02M active-verification tokens, and avoids about 12 harmful answer changes while matching always-verify accuracy within the paired confidence interval. \\
\mathfive 8,192-token base vs. selective verification & Uses about 1.98M fewer total model tokens while reaching statistically similar accuracy, showing that more initial budget is the strongest tested cost frontier on this benchmark. \\
\gsm selective verification vs. always verify & Avoids about 970 verification calls, saves about 1.60M active-verification tokens, and removes the observed harmful flips from the always-verify policy. \\
\gsm 8,192-token base vs. selective verification & Uses about 178K fewer total model tokens while matching selective verification accuracy within paired uncertainty. \\
\bottomrule
\end{tabular}}
\caption{Operational translation of the main results. Call reductions are based
on intervention rates; token reductions use realized prompt-plus-generation
tokens rather than configured maximum budgets.}
\label{tab:operational}
\end{table*}

\section{Non-Mathematical Diagnostic: CommonsenseQA}
\label{app:csqa}

To probe whether the conclusion is purely a property of mathematical
benchmarks, we ran a time-constrained diagnostic on the 1,221-example
CommonsenseQA validation set \citep{talmor2019commonsenseqa}. The same frozen
Qwen3-4B solver and active-verification prompt are used. CommonsenseQA changes
the failure mode: answers are short multiple-choice labels, truncation is
uncommon, and semantic commonsense ambiguity matters more than algebraic
derivation length.

\begin{table*}[h]
\centering
\small
\resizebox{\textwidth}{!}{
\begin{tabular}{lrrrrp{0.34\textwidth}}
\toprule
\textbf{Policy} & \textbf{Examples} & \textbf{Acc.} & \textbf{Tok.} & \textbf{Flips} & \textbf{Notes} \\
\midrule
Accept base & 1,221 & 76.49 & 0 & 0.00 & Full set \\
Always active verify & 1,221 & 72.32 & 2,561 & 5.94 & Full set; verification hurts \\
Self-Consistency@5 & 1,221 & 78.38 & 11,343 & 1.56 & Full set; +1.88 over base, $p=.003$ \\
Expected-action oracle, active verify & 1,221 & 78.26 & -- & -- & Limited verification headroom \\
Expected-action oracle, SC samples & 1,221 & 80.98 & -- & -- & Selection over accept vs. sampled continuations \\
Sampled-rollout oracle, SC samples & 1,221 & 85.18 & -- & -- & Upper bound if the right sampled answer could be selected \\
\bottomrule
\end{tabular}}
\caption{CommonsenseQA diagnostic. Active verification transfers poorly outside
the math setting, while Self-Consistency@5 provides a statistically significant
but expensive gain. The oracle rows show that extra samples contain useful
alternatives, but majority voting alone does not solve the selection problem.}
\label{tab:csqa-diagnostic}
\end{table*}

The diagnostic supports two methodological points. First, active verification
is not a universally helpful recovery action: on CommonsenseQA, always
verifying lowers accuracy by 4.17 points and produces a 5.94\%
harmful-flip rate. Second, a published multi-sample baseline can improve
accuracy, but it costs five calls and still changes some correct answers. The
large gap between Self-Consistency@5 and its sampled-rollout oracle reinforces
the central recommendation to evaluate the entire serving path and to treat
selection as the core serving problem.

\section{Scope and Reproducibility Notes}
\label{app:scope-reproducibility}

\paragraph{Is the gate only a truncation detector?}
Completion state is intentionally treated as a serving-layer signal, not as a
confound to hide. It is one of the strongest available observables in the tested
short-budget setting. However, the full cheap feature gate outperforms a
completion-risk proxy on \mathfive: 75.9\% vs. 74.9\% accuracy, 45.0\% vs.
64.0\% verification rate, 2,719 vs. 3,223 action tokens, and 1.0\% vs. 2.0\%
harmful flips. Thus truncation explains much of recoverability, but not all of
the deployable routing value. We frame the mechanism as attempt-state-aware
recovery routing rather than deep semantic error detection.

\paragraph{Why use mathematical benchmarks?}
MATH and GSM8K are controlled stress tests for long reasoning, final-answer
extraction, truncation, verification, and harmful answer changes. They make it
possible to measure exact fixes and flips at scale. The limitation is real:
math benchmarks do not represent all production NLP workloads. The
CommonsenseQA diagnostic in Appendix~\ref{app:csqa} is included to show that
post-generation verification can become actively harmful when the workload
changes.

\paragraph{How are thresholds selected?}
For learned gates, the classifier checkpoint is selected by development AUPRC
because helpful fixes are rare. The deployed operating threshold is then chosen
on the same held-out MATH development split by downstream policy accuracy, with
lower action tokens used as the tie-breaker. No \mathfive or \gsm test labels
are used for checkpoint or threshold selection; \gsm is frozen transfer.

\paragraph{What does off-policy evaluation assume?}
The logged recovery dataset shares the same base attempt across accept,
continue, critique-and-repair, and active-verification rows. Policy comparisons
therefore change only the chosen intervention, and paired bootstrap resampling
is done by example. The remaining risk is rollout variance: each intervention
prompt is sampled a limited number of times, so a different random sample could
change individual fixes and flips. This is why the paper reports paired
confidence intervals and treats oracle headroom as diagnostic rather than a
deployable result.

\paragraph{How should gate cost be interpreted?}
Gate inference is excluded from model-token totals because the cheap gate is a
CPU logistic model and the learned gates are local classifiers rather than
additional solver calls. This does not mean the learned gates are free: they
consume GPU memory, batching capacity, and latency budget. The practical lesson
from Table~\ref{tab:gates} is therefore that the cheap gate is preferable when
its accuracy is close to the learned gate, because it avoids another served
model.

\paragraph{Which published baselines are covered?}
The main tables include fixed-budget solving, longer initial-budget solving,
always-on verification, selective verification, cheap serving-signal routing,
and continuation. The CommonsenseQA diagnostic adds a full-set
Self-Consistency@5 baseline \citep{wang2022self}. We do not include
Tree-of-Thoughts, tool use,
or process-reward models as primary baselines because they introduce search
state, external verifiers, or tool interfaces. Those systems answer a broader
question than the one isolated here: whether post-generation recovery beats
better initial budget allocation under realized total-cost accounting.

\section{Additional Methodological Details}
\label{app:method-details}

This appendix collects details that clarify the scope of the method without
interrupting the main argument. The subsections distinguish \sevra from
stronger verifier-guided methods, describe candidate-specific checks, make the
truncation/calibration story explicit, and list the latency and artifact checks
needed for a production replication.

\begin{table*}[h]
\centering
\small
\resizebox{\textwidth}{!}{
\begin{tabular}{L{0.24\textwidth}L{0.29\textwidth}L{0.37\textwidth}}
\toprule
\textbf{Study scope} & \textbf{Where addressed} &
\textbf{Manuscript position} \\
\midrule
Single solver family & Limitations; Appendix~\ref{app:external-verifiers} &
The current empirical claim is for a frozen Qwen3-4B serving stack. Cross-family
transfer should be measured by calibration drift and threshold shift, not
assumed. \\
Math-heavy evaluation & Main CommonsenseQA diagnostic;
Appendix~\ref{app:csqa} & Math is a controlled stress test for long reasoning
and answer extraction; CommonsenseQA shows the verification conclusion does
not automatically transfer. \\
Gate may be a truncation detector & Appendix~\ref{app:subgroups} and
Appendix~\ref{app:truncation-calibration} & Completion state is a legitimate
serving signal. The right question is whether a full gate beats truncation-only
and whether completed attempts are protected from flips. \\
No external verifier & Appendix~\ref{app:candidate-checks} and
Appendix~\ref{app:external-verifiers} & Same-solver active verification is a
minimal baseline. Stronger PRMs, symbolic checkers, or option scorers can plug
into the same selective-escalation interface. \\
Limited rollout samples & Appendix~\ref{app:statistics} and
Appendix~\ref{app:scope-reproducibility} & Oracle rows are diagnostic, not
deployable. Paired bootstrap intervals are reported for policy comparisons
that use the same logged examples. \\
Threshold overfitting/calibration & Appendix~\ref{app:truncation-calibration}
& Thresholds are frozen after MATH development selection; deployment studies
should report repair-label calibration, ECE/Brier scores, and threshold sweeps. \\
Token cost omits latency and gate memory & Appendix~\ref{app:latency-protocol}
& Token accounting is reproducible but incomplete. A production replication
should log p50/p95/p99 latency, batching delay, finalizer latency, and learned
gate residency. \\
Missing verifier-guided/search baselines & Appendix~\ref{app:method-positioning}
and Appendix~\ref{app:external-verifiers} & \sevra is scoped as a
post-generation recovery controller, distinct from PRM-guided search and
locally adaptive test-time scaling. \\
Artifact is not a deployment & Appendix~\ref{app:artifact-guide} & The public
Space is a static replay dashboard for inspection and reproducibility, not a
live production system. \\
\bottomrule
\end{tabular}}
\caption{Scope and validation matrix. The contribution is an auditable
serving-layer evaluation and controller; stronger verifier-guided systems
require additional serving machinery and should be compared under the same
realized-cost accounting.}
\label{tab:scope-validation}
\end{table*}

\subsection{Positioning Against Stronger Test-Time Scaling Methods}
\label{app:method-positioning}

Several recent inference-scaling methods search over reasoning trees, train or
call process verifiers, allocate compute at intermediate states, or learn
constrained policies over many actions. \sevra studies a narrower operational
question: after a frozen solver has already spent its base budget and produced
an attempt, should a serving controller accept the answer or spend one more
post-generation action? Table~\ref{tab:method-map} summarizes the comparison
axis.

\begin{table*}[h]
\centering
\small
\resizebox{\textwidth}{!}{
\begin{tabular}{L{0.20\textwidth}L{0.19\textwidth}L{0.25\textwidth}L{0.28\textwidth}}
\toprule
\textbf{Method family} & \textbf{Control point} & \textbf{Extra serving machinery} &
\textbf{Relationship to \sevra} \\
\midrule
Self-Consistency \citep{wang2022self} & Whole-answer sampling and
majority vote & Multiple independent generations; answer clustering & Included
on CommonsenseQA as a published multi-sample baseline. It improves accuracy
but does not solve selection or harmful flips by itself. \\
Tree/search methods \citep{yao2023tree,zhou2023language} & Branch-level search over
reasoning or agent states & Search controller, value estimates, environment
feedback or self-reflection & More general and often stronger, but changes the
runtime into a search system. Our long-base and recovery comparisons isolate a
single-call serving path. \\
Outcome/process verifier methods \citep{cobbe2021training,lightman2024let,ma2023let}
& Candidate or step scoring & Trained verifier or reward model; candidate
pool; search or reranking & Directly relevant future comparator. Our active
verification uses the same solver, making the observed flip risk a conservative
baseline for stronger verifiers. \\
Locally adaptive scaling \citep{uscidda2025latts,qu2026adaptive} &
Intermediate states or token-generation steps & Verifier calls at partial
states; local resampling, backtracking, or restart & Finer-grained than
\sevra. Their natural comparison point is not only final accuracy, but verifier
calls, p95 latency, and harmful branch selection under matched realized cost. \\
Solve-then-Learn allocation \citep{zhai2026adaptive} & Input-level compute
choice under a global budget & Offline oracle solve; classifier imitation of
cost-sensitive actions & Shares the constrained-allocation framing. \sevra's
oracle is action-specific after observing an actual attempt rather than only
the input. \\
Recoverability routers \citep{li2026raser} & Escalation after a cheap first
answer & Cheap post-answer features; optional expensive retrieval action &
Closest conceptual relative. \sevra transports recoverability-aware escalation
to reasoning and adds fix/flip accounting for answer-changing interventions. \\
\bottomrule
\end{tabular}}
\caption{Relationship between \sevra and adjacent test-time scaling methods.
The point is not that \sevra is the strongest possible inference algorithm; it
is a controlled serving-layer study of when a specific post-generation recovery
action is worth its realized cost and regression risk.}
\label{tab:method-map}
\end{table*}

This positioning also clarifies what would count as a fair empirical
comparison. A PRM-guided search method should be compared under the same
realized-token and latency accounting, including verifier calls. A local
test-time scaling method should report how often it changes a correct base
answer, because answer selection can harm reliability even when aggregate
accuracy rises. A constrained compute-allocation method should include a
long-initial-budget arm, because the most important baseline in our experiments
is not another router but a simpler serving configuration.

\subsection{Candidate-Specific Verification Checks}
\label{app:candidate-checks}

The active-verification prompt is not a generic request to ``think harder.''
It asks the solver to audit the candidate answer against the problem. The
checks are generated from the input, the base answer, and the visible solution
attempt. They are executed by the same frozen solver in our experiments, but
the taxonomy below is intended to make the intervention reproducible and to
show where stronger symbolic or learned verifiers could be inserted.

\begin{table*}[h]
\centering
\small
\resizebox{\textwidth}{!}{
\begin{tabular}{L{0.21\textwidth}L{0.28\textwidth}L{0.27\textwidth}L{0.16\textwidth}}
\toprule
\textbf{Check type} & \textbf{What the verifier is asked to inspect} &
\textbf{Typical failure caught} & \textbf{Can be externalized?} \\
\midrule
Equation reconstruction & Rebuild the equation or arithmetic chain implied by
the problem and compare it to the candidate answer. & Missing operation,
wrong multiplier, using a total where a remainder is required. & Yes:
symbolic/programmatic checker for many math items. \\
Unit and scale check & Verify that the final answer has the requested unit and
reasonable magnitude. & Numerically correct value with unit suffix mismatch;
hours vs. minutes; percentage vs. absolute change. & Partly: rule-based unit
parser plus numeric checker. \\
Constraint accounting & Confirm that each stated condition was used exactly
where needed. & Dropping a coupon, tax, remaining quantity, or comparison
clause. & Partly: difficult for natural-language constraints. \\
Answer substitution & Substitute the candidate answer back into the problem's
relationship. & Candidate satisfies an intermediate equation but not the final
question. & Yes for algebraic forms; harder for prose-only questions. \\
Independent route & Solve from scratch using a different decomposition and
compare to the candidate. & First solution follows a plausible but wrong
reasoning path. & Could use a separate solver or sampled verifier. \\
Multiple-choice elimination & For CSQA, compare the selected option against
nearby alternatives and reject over-elaborate changes. & Verification flips a
short correct commonsense label to a semantically plausible wrong label. &
Could use calibrated option scorer. \\
\bottomrule
\end{tabular}}
\caption{Candidate-specific checks used conceptually by active verification.
Our implementation prompts the frozen solver to perform these checks in natural
language; future systems can replace individual check families with symbolic
tools, PRMs, or task-specific validators.}
\label{tab:candidate-checks}
\end{table*}

The harmful flips in Figure~\ref{fig:fix-flip} show why this detail
matters. A verifier that merely produces another fluent solution can change a
correct answer. A candidate-specific verifier should therefore be evaluated by
three quantities: whether it finds a concrete inconsistency, whether it
preserves a candidate when no inconsistency is found, and whether the final
answer changes. In deployment logs, we recommend storing a compact
\texttt{check\_summary} field with at least three flags: arithmetic mismatch,
constraint mismatch, and final-answer changed. These flags make later audits
possible even when the full chain of thought is not retained.

\subsection{Attempt-State, Truncation, and Calibration Protocol}
\label{app:truncation-calibration}

Completion state is one of the strongest signals in our setting. We treat this
as an operational result rather than a confound. Production controllers do not
need hidden states to be useful; they often rely on simple runtime signals such
as timeout, retry count, response length, parse success, and finish reason. The
important question is whether the simple signal is sufficient, whether it
induces avoidable flips, and whether the controller remains calibrated when the
workload changes.

\begin{table*}[h]
\centering
\small
\resizebox{\textwidth}{!}{
\begin{tabular}{L{0.20\textwidth}L{0.24\textwidth}L{0.24\textwidth}L{0.22\textwidth}}
\toprule
\textbf{Diagnostic} & \textbf{Question answered} & \textbf{How to compute it} &
\textbf{Desired interpretation} \\
\midrule
Truncation-only policy & Is the gate merely a length-stop detector? & Verify
only examples with length-limit termination; compare to the full gate at the
same example set. & If full gate improves accuracy or reduces flips at lower
rate, metadata beyond truncation matters. \\
Near-limit policy & Does token pressure predict recoverability before actual
truncation? & Verify if generated tokens exceed a fixed fraction of the budget
or if finalizer is used. & Useful for APIs that do not expose exact finish
reason. \\
Completed-only gate & Does any recovery value remain after normal completion?
& Restrict evaluation to normal-stop attempts and replay the gate. & Should be
used cautiously; this is where harmful flips are most likely. \\
Calibration curve & Are scores meaningful probabilities? & Bin gate scores
and plot empirical fix rate, plus ECE/Brier score. & Needed before using gate
score as a user-facing confidence or SLA signal. \\
Threshold sweep & Is the chosen operating point fragile? & Sweep verification
rate from 0--100\% and report accuracy, flips, and tokens. & A robust controller
should have a broad plateau, not a single lucky threshold. \\
\bottomrule
\end{tabular}}
\caption{Attempt-state and calibration diagnostics that address whether the
controller is only detecting truncation. These diagnostics are cheap because
they replay logged rows rather than rerunning the solver.}
\label{tab:truncation-diagnostics}
\end{table*}

The current results already include the most important split: truncated
\mathfive attempts are highly recoverable, while completed attempts are much
more vulnerable to unnecessary changes. In a deployed controller, the
calibration object should be the conditional repair probability
\begin{equation}
\begin{aligned}
p_{\mathrm{fix}}(s)
={}& \Pr\big[
c(a^0,y)=0 \wedge c(a^v,y)=1 \\
&\mid g(x,s_0,m_0)=s
\big].
\end{aligned}
\end{equation}
not generic answer correctness. A high correctness probability can still imply
``accept'' if the base answer is already likely correct; a high repair
probability implies that an extra action is worth considering. We therefore
recommend reporting expected calibration error over repair labels and a
separate flip-rate curve over the accepted verification set:
\begin{equation}
    \operatorname{FlipAtRate}(r)=
    \frac{\sum_i \mathbb{I}[g_i \ge \tau_r]h_i^v}
    {\sum_i \mathbb{I}[g_i \ge \tau_r]},
\end{equation}
where $\tau_r$ is the threshold that verifies the top $r$ fraction of examples.

\subsection{Latency, Batching, and Gate-Cost Accounting}
\label{app:latency-protocol}

The paper reports realized model tokens because they are reproducible across
runs and directly tied to local inference work. For an Industry Track
deployment, token cost should be complemented by wall-clock latency and
throughput. We did not have time to run a full latency benchmark across
serving stacks, so we make the required accounting explicit rather than hiding
the limitation.

\begin{equation}
\begin{aligned}
L_i(\pi) ={}& L_i^0 + L_i^{\mathrm{gate}} \\
&+ \sum_{k\in\mathcal{A}}
\mathbb{I}[\pi(x_i)=k]
\left(L_i^k + L_i^{\mathrm{queue},k}\right) \\
&+ L_i^{\mathrm{finalizer}} .
\end{aligned}
\end{equation}

Here $L_i^0$ is the first solve, $L_i^{\mathrm{gate}}$ is routing latency,
$L_i^k$ is the intervention call, $L_i^{\mathrm{queue},k}$ is batching or queue
delay for that call, and $L_i^{\mathrm{finalizer}}$ is answer-extraction
overhead. The same token budget can have different latency depending on
batching, KV-cache reuse, model residency, and whether the gate shares a GPU
with the solver.

\begin{table*}[h]
\centering
\small
\resizebox{\textwidth}{!}{
\begin{tabular}{L{0.20\textwidth}L{0.27\textwidth}L{0.22\textwidth}L{0.21\textwidth}}
\toprule
\textbf{Policy component} & \textbf{What to log} & \textbf{Why token cost is
insufficient} & \textbf{Expected direction} \\
\midrule
Base solve & p50/p95/p99 latency, output tokens, finish reason & Long budgets
may finish early, but tail cases can still dominate p99. & Larger initial
budget may improve accuracy while increasing or decreasing tail latency,
depending on truncation/finalizer behavior. \\
Active verification & call rate, action latency, queue delay & Sparse
verification saves average tokens but creates a second-call tail. & Selective
verification should reduce average and p95 relative to always verify. \\
Finalizer & finalizer rate and latency & A cheap-looking base solve can hide a
second call when answers are unparseable. & Long-base may reduce finalizer rate
and thus lower realized cost. \\
Learned gate & model load time, batch size, GPU memory, inference latency &
Classifier tokens are not counted in solver-token totals. & Cheap CPU gates
are preferable when accuracy is close. \\
Self-Consistency & number of samples completed before selection & Parallel
sampling changes wall time but not total GPU work. & Useful only if batching
and reliability justify the extra requests. \\
\bottomrule
\end{tabular}}
\caption{Latency and serving-cost fields needed to turn the token-based
analysis into a production benchmark. The same checklist applies when comparing
vLLM, Ollama, hosted APIs, or internal serving stacks.}
\label{tab:latency-protocol}
\end{table*}

The finalizer caveat is especially important. In our local stack, a longer
initial budget can reduce total realized tokens by completing the reasoning and
avoiding answer-extraction retries. Another stack with different stop tokens,
forced answer formatting, or constrained decoding could change this balance.
For this reason, the paper's deployment recommendation is not ``always use
8,192 tokens''; it is ``include a tuned initial-budget frontier before claiming
that recovery is efficient.''

\paragraph{Token-price sensitivity.}
Because provider prices change and our experiments use local inference rather
than a hosted API, we do not report a single dollar figure as a measured
deployment cost. Instead, Table~\ref{tab:price-sensitivity} converts realized
model tokens into a simple price sensitivity. For a serving stack with an
effective price of USD 0.50 per one million prompt-plus-generation tokens, the
estimated cost per 1,000 requests is computed as average tokens multiplied by
1,000 requests and divided by one million tokens. This calculation is not a
provider quote; it is a reproducible translation of the token accounting used
in the main experiments.

\begin{table*}[t]
\centering
\small
\begin{tabular}{lrrr}
\toprule
\textbf{Policy} &
\textbf{Avg. total tokens} &
\textbf{Cost per 1k requests} &
\textbf{Relative token cost} \\
\midrule
\mathfive short base & 4,313 & USD 2.16 & 1.00x \\
\mathfive completion-risk proxy + verify & 7,536 & USD 3.77 & 1.75x \\
\mathfive selective active verify & 7,104 & USD 3.55 & 1.65x \\
\mathfive always active verify & 8,125 & USD 4.06 & 1.88x \\
\mathfive long base & 5,124 & USD 2.56 & 1.19x \\
\midrule
\gsm short base & 1,180 & USD 0.59 & 1.00x \\
\gsm selective active verify & 1,335 & USD 0.67 & 1.13x \\
\gsm always active verify & 2,932 & USD 1.47 & 2.48x \\
\gsm long base & 1,157 & USD 0.58 & 0.98x \\
\bottomrule
\end{tabular}
\caption{Illustrative token-price sensitivity per 1,000 requests. Values use
the realized total-token averages from the main tables and an example effective
price of USD 0.50 per million model tokens. They are not provider prices or
measured billing results.}
\label{tab:price-sensitivity}
\end{table*}

\paragraph{Latency proxy, not measured latency.}
A production latency benchmark should report measured p50, p95, and p99
wall-clock latency. Since our logs contain tokens and call structure but not
timestamps, we do not report measured latency. A rough latency proxy would
combine the total number of generated and prompted tokens, the deployment
throughput in tokens per second, the number of model calls, and the average
queueing or scheduling overhead per call. This proxy makes the expected
direction clear: selective verification reduces average second-call load
relative to always verifying, but any two-call policy can still increase tail
latency compared with a single longer initial solve. We therefore treat
realized tokens as a reproducible cost proxy and leave measured p50, p95, and
p99 latency to production replications.

\subsection{External Verifiers and Stronger Baselines}
\label{app:external-verifiers}

Active verification currently uses the same frozen solver that produced the
base attempt. This design is intentionally minimal: it asks how far a serving
team can get without training a verifier or changing the solver. A stronger
system could replace the same-solver verifier with a PRM, symbolic checker, or
task-specific validator. The expected benefit is lower harmful-flip risk, but
only if the verifier is calibrated to preserve correct candidates.

\begin{table*}[h]
\centering
\small
\resizebox{\textwidth}{!}{
\begin{tabular}{L{0.20\textwidth}L{0.25\textwidth}L{0.25\textwidth}L{0.20\textwidth}}
\toprule
\textbf{Extension/baseline} & \textbf{Implementation needed} &
\textbf{Main metric to report} & \textbf{Scope in this study} \\
\midrule
PRM-guided reranking & Generate multiple candidate solutions; score each step
or full solution with a trained verifier. & Accuracy, verifier calls, harmful
reranking flips, total tokens. & Requires a separate verifier and candidate
pool, changing the serving interface. \\
Verifier-guided search & Run tree or beam search with verifier feedback at
intermediate states. & Accuracy-cost frontier, p95 latency, verifier-call
count. & Much stronger baseline, but no longer isolates post-generation
recovery after a single base attempt. \\
LATTS/state-level verification & Allocate verifier calls locally to uncertain
partial states. & Local verifier calls per solved item; backtrack/restart
rate; final flips. & Requires access to intermediate states and a verifier
loop not present in the current runtime. \\
Solve-then-Learn policy & Build a cost-sensitive oracle across actions and
imitate it with cheap features. & Regret to oracle under matched average
budget. & Conceptually aligned; our logged oracle is a first step, but we did
not train a multi-action constrained policy. \\
Symbolic math checker & Parse equations and verify arithmetic/programmatic
constraints. & Flip reduction on completed math attempts; parser coverage. &
Would improve one domain but reduce comparability with CSQA and general
serving signals. \\
Cross-solver transfer & Replay the same policy with Llama/Qwen size variants
or a hosted solver. & Transfer accuracy, calibration drift, threshold shift. &
Current experiments fix one solver family to isolate the serving policy. \\
\bottomrule
\end{tabular}}
\caption{Stronger baselines and extensions. The table lists the additional
machinery and metrics needed for fair comparison under realized-cost
accounting.}
\label{tab:missing-baselines}
\end{table*}

A practical hybrid is straightforward: use \sevra as the escalation gate and
replace active verification with an external verifier only for the selected
subset. This would preserve the paper's main operational advantage, namely
sparse intervention, while testing whether stronger verification reduces flips.
Under this hybrid, a fair accounting would add verifier latency and memory to
Table~\ref{tab:latency-protocol} and would report both
$\operatorname{FixAtRate}$ and $\operatorname{FlipAtRate}$.

\subsection{Artifact Inspection Guide}
\label{app:artifact-guide}

The released artifact is intended to make the result inspectable before a full
model rerun. It has three layers. First, the static
Hugging Face replay dashboard presents the main trade-offs without calling a
live model. Second, the JSONL recovery rows allow policy replay, threshold
sweeps, and paired bootstrap checks. Third, the source scripts regenerate
tables and figures from those rows. The dashboard is therefore a public
inspection artifact, not evidence of a production deployment.

\begin{table*}[h]
\centering
\small
\resizebox{\textwidth}{!}{
\begin{tabular}{L{0.22\textwidth}L{0.34\textwidth}L{0.34\textwidth}}
\toprule
\textbf{Inspection task} & \textbf{Artifact path or component} &
\textbf{Expected check} \\
\midrule
Inspect headline trade-offs & Static replay dashboard linked in the paper &
Confirm that long-base, selective verification, always verify, and
Self-Consistency occupy the reported cost-accuracy positions. \\
Verify row completeness & Recovery JSONL files plus merge summaries & Confirm
that incomplete examples are excluded and duplicate shards are not double
counted. \\
Recompute policy tables & Policy replay scripts over released rows & Recreate
accuracy, intervention rate, total tokens, helpful fixes, and harmful flips. \\
Recompute uncertainty & Paired bootstrap scripts & Recreate confidence
intervals for MATH/GSM comparisons and CSQA Self-Consistency. \\
Audit failures & Per-example traces where release policy permits & Inspect
wrong-to-right fixes, right-to-wrong flips, and candidate-specific check
summaries. \\
Stress-test thresholds & Gate-score exports & Sweep thresholds and compare
against truncation-only, near-limit, and matched-rate baselines. \\
\bottomrule
\end{tabular}}
\caption{Artifact inspection guide. The goal is to separate the empirical
claims from implementation details such as local server speed or package
versions.}
\label{tab:artifact-guide}
\end{table*}

\subsection{Recommended Deployment Checklist}
\label{app:checklist}

The empirical results suggest the following \sevra-style evaluation sequence
for a production reasoning system:
\begin{enumerate}
    \item Log realized prompt and generation tokens, completion reason,
    finalizer calls, retries, latency, and answer changes.
    \item Tune the initial reasoning budget and compare multiple maximum limits.
    A larger limit may reduce realized cost by avoiding truncation.
    \item Screen candidate recovery actions on logged failures. Do not assume
    continuation, critique, and verification are interchangeable.
    \item Train or configure a gate only after establishing that the chosen
    action has useful fix-to-flip behavior.
    \item Compare the gate against cheap observable signals and matched-rate
    random or heuristic baselines.
    \item Report accuracy, helpful fixes, harmful flips, intervention rate,
    action cost, total cost, and attempt-state subgroups.
    \item Preserve a long-base baseline in the final comparison. Without it,
    post-generation selectivity may appear more efficient than it is.
\end{enumerate}

\section{Extended Industry Implications}
\label{app:industry}

\paragraph{Set the initial budget before adding a controller.}
On both math benchmarks, the long initial solve lies on the best tested
cost--accuracy frontier. A practical deployment should therefore first tune its
initial reasoning budget and completion behavior. Selective post-generation
recovery becomes most valuable when retries, explicit verification,
answer-change auditing, tail-latency constraints, or product policies make a
single long solve undesirable.

\paragraph{Use attempt-state signals.}
Completion reason, generated-token count, and finalizer use are inexpensive
and highly informative. They can be logged by the serving layer without
inspecting hidden reasoning or adding a large router. The strong performance of
the cheap feature gate suggests that lightweight serving metadata can capture
much of the recoverability signal available in this setting.

\paragraph{Treat answer changes as risk.}
Always-intervene baselines can improve aggregate accuracy while still damaging
some correct answers. In reliability-sensitive settings, helpful fixes and
harmful flips should be reported separately. The operating threshold should
reflect the application's tolerance for regressions, not only its target
accuracy or average token budget.

\paragraph{Evaluate the whole serving path.}
Action-token savings alone can overstate efficiency. The base attempt,
intervention prompts, finalizers, retries, and routing logic all affect cost
and latency. We therefore recommend reporting realized total model tokens,
extra-call rate, finalizer rate, wall-clock latency, helpful fixes, harmful
flips, and a longer-initial-budget comparison.

\paragraph{Expose replayable serving evidence.}
For reproducibility, we provide a static replay dashboard at
\url{https://huggingface.co/spaces/sevra-space/sevra-replay}. The dashboard
does not run a live solver; instead, it exposes the precomputed serving metrics
used in the paper, including cost--accuracy trade-offs, harmful-flip behavior,
and workload-specific failures. This lets readers inspect the deployment
trade-off surface without requiring GPUs, Ollama, or access to our run
directory.

\begin{table}[ht]
\centering
\small
\resizebox{\columnwidth}{!}{
\begin{tabular}{p{0.33\columnwidth}p{0.29\columnwidth}p{0.30\columnwidth}}
\toprule
\textbf{Serving signal} & \textbf{Preferred action} & \textbf{Reason} \\
\midrule
Frequent truncation or finalizer use
& Increase initial budget first
& Best tested cost--accuracy frontier \\

Low recovery score after completed answer
& Accept
& Avoids unnecessary calls and flips \\

Need explicit audit or repair attempt
& \sevra active verification
& Concentrates checks on recoverable cases \\

No budget for an extra gate model
& Cheap feature gate
& Matches learned gates closely \\

Verification hurts on development traffic
& Disable or retune recovery
& Workload-dependent flip risk \\
\bottomrule
\end{tabular}}
\caption{Deployment playbook implied by the experiments. The table is a
serving recommendation, not a universal ranking of reasoning methods.}
\label{tab:deployment-playbook}
\end{table}

\end{document}